\documentclass[11pt]{article}

\usepackage[preprint]{acl}

\usepackage{times}
\usepackage{latexsym}

\usepackage[T1]{fontenc}

\usepackage[utf8]{inputenc}

\usepackage{enumitem}

\usepackage{microtype}

\usepackage{inconsolata}

\usepackage{graphicx}

\usepackage{hyperref}
\usepackage{url}
\usepackage{booktabs}
\usepackage{url}
\usepackage{wrapfig}
\usepackage{graphicx}
\usepackage{multirow}
\usepackage{subcaption}
\usepackage{multicol}

\usepackage{algorithm}
\usepackage{algpseudocode}
\usepackage{xcolor}
\usepackage{pifont}
\usepackage{makecell}
\usepackage[dvipsnames]{xcolor} 
\usepackage{stackengine}

\newcommand{\q}[1]{``#1''} 

\definecolor{eblue}{rgb}{0.36, 0.36, 0.36}
\definecolor{limegreen}{rgb}{0.2, 0.8, 0.2}
\usepackage{cleveref}
\usepackage{tcolorbox}

\newtcolorbox[auto counter, number within=section]{obsbox}[2][]{%
  colback=blue!5!white,
  colframe=blue!75!black,
  fonttitle=\bfseries,
  title=Observation~\thetcbcounter: #2,
  label=#1 
}
\crefname{obsbox}{Observation}{Observations}

\definecolor{gemGreen}{RGB}{34, 139, 34}
\definecolor{gemRed}{RGB}{200, 50, 50}
\definecolor{gemOrange}{RGB}{255, 140, 0}

\newcommand{\docYes}{\textcolor{gemGreen}{\ding{51}}}   
\newcommand{\docNo}{\textcolor{gemRed}{\ding{55}}}   
\newcommand{\docIrr}{\textcolor{gemOrange}{\ding{108}}} 

\title{Discovering Hidden Gems in Model Repositories
}

\author{Jonathan Kahana \and Eliahu Horwitz \and Yedid Hoshen \\
Department of Computer Science\\
School of Computer Science and Engineering\\
The Hebrew University of Jerusalem, Israel\\
\texttt{jonathan.kahana@mail.huji.ac.il} \\
\tt\href{https://jonkahana.github.io/hidden_gems/}{\textbf{https://jonkahana.github.io/hidden\_gems}}
}

\begin{document}

\maketitle

\begin{abstract}
Public repositories host millions of fine-tuned models, yet community usage remains disproportionately concentrated on a small number of foundation checkpoints. We investigate whether this concentration reflects efficient market selection or if superior models are systematically overlooked. Through an extensive evaluation of over $2,000$ models, we show the prevalence of \q{hidden gems}, unpopular fine-tunes that significantly outperform their popular counterparts. Notably, within the Llama-3.1-8B family, we find rarely downloaded checkpoints that improve math performance from $83.2\%$ to $96.0\%$ without increasing inference costs. However, discovering these models through exhaustive evaluation of every uploaded model is computationally infeasible. We therefore formulate model discovery as a Multi-Armed Bandit problem and accelerate the Sequential Halving search algorithm by using shared query sets and aggressive elimination schedules. Our method retrieves top models with as few as $50$ queries per candidate, accelerating discovery by over $50\times$.
\end{abstract}

\section{Introduction}

Public model repositories, such as Hugging Face, have democratized access to new, high-performing models \citep{casper2025open}. However, as these repositories scale to millions of checkpoints, selecting the right model for each task becomes a critical bottleneck. Relying on documentation is often ineffective, as model cards \citep{model_cards} are frequently incomplete or missing entirely \citep{kahana2025can,horwitz2025we}. Consequently, the vast majority of users default to the foundation model at the root of the Model Tree \citep{Horwitz_2025_CVPR}, such as the official Qwen or Llama checkpoints or to its official instruction fine-tuning. While this strategy is simple, its efficacy is unproven. This paper tackles two questions: i) Are the most popular models actually the best performers available? ii) If not, how can we efficiently identify superior models among many candidates?

\begin{figure}[t]
    \centering
    \includegraphics[width=\linewidth]{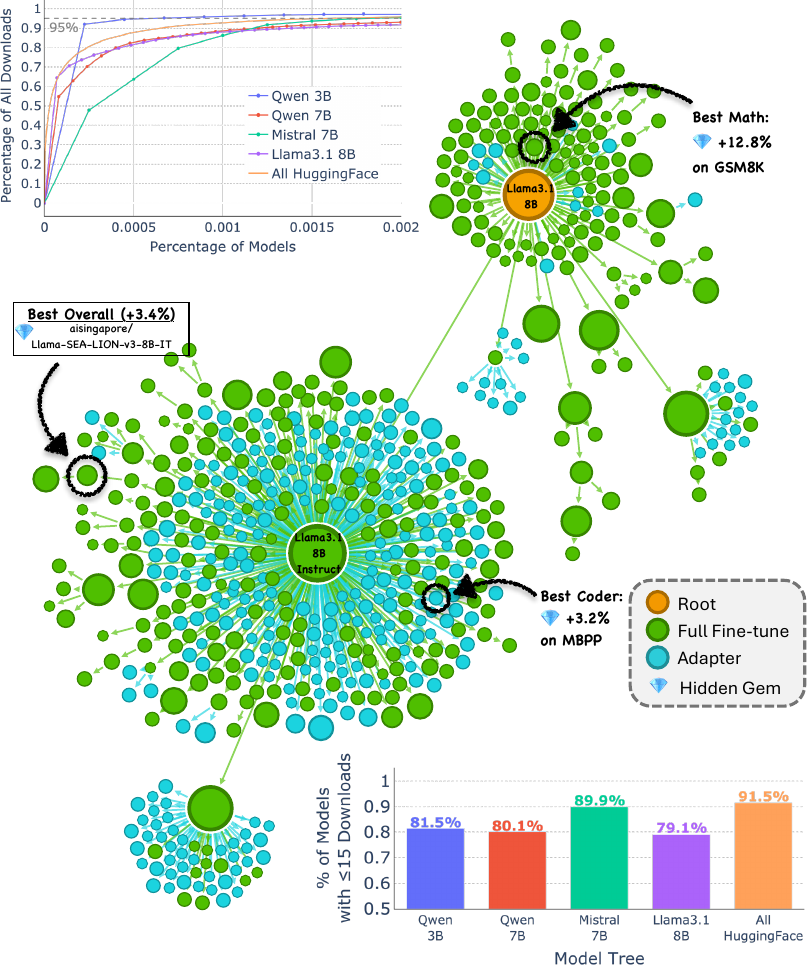}
    \caption{\textbf{\textit{Hidden Gems \& Repository Inefficiency.}} \textbf{Center:} The Llama3.1-8B Model Tree, where node size reflects downloads (log scale). Our evaluation reveals \q{hidden gems} (circled): unpopular models that significantly outperform widely used baselines. \textbf{Top Left:} Cumulative download rates showing usage is extremely concentrated in a tiny fraction of top models. \textbf{Bottom Right:} The vast majority of models (over $90\%$) are rarely explored, receiving $\leq15$ monthly downloads.}
    \label{fig:usability}
    \vspace{-0.65cm}
\end{figure}

First, we investigate whether users are successful at identifying the best models by analyzing the relation between download counts and performance. By evaluating more than $2,000$ models from several popular model families on diverse tasks, we consistently identify unpopular models that significantly outperform their popular counterparts. We term these lost high-performers \q{hidden gems}.

Discovering such hidden gems presents a significant challenge. Exhaustively evaluating all candidate models may require billions of inferences. Unlike existing methods that rank new models within a predefined leaderboard \citep{perlitz-etal-2024-efficient,tamura2025can,ashury2024label}, we aim to rank an entire model population from scratch without knowing the rank of previous models. 
Our approach can be easily integrated with query selection methods \citep{polo2024tinybenchmarks,zouhar2025selectdatapointsefficienthuman} although we do not focus on this here. Concretely, we formulate model discovery as a Multi-Armed Bandit (MAB) problem \citep{katehakis1987multi}. We adapt the Sequential Halving algorithm \citep{karnin2013almost} to exploit the unique characteristics of model evaluation, such as the ability to share query sets across models for variance reduction. Our approach consistently finds a top-$3$ model with just $50$ queries per model, making the search over $50\times$ faster than exhaustive baselines, while improving average performance by over $4.5\%$.

\setlength{\tabcolsep}{3pt} 
\begin{table}[t]
    \caption{\textbf{\textit{Model Discovery Results.}} Comparison of popular (Base) and gem models performance. For each model tree, we present the base performance, the discovered gem performance, and the resulting gain. RouterBench$_s$ is the combined performance of all other tasks.}
    \centering
    \resizebox{\linewidth}{!}{
        \large 
        \begin{tabular}{lccccccc}
            \textbf{Tree} &  & \textbf{ARC-C$_s$} & \textbf{WinoG.$_s$} & \textbf{MMLU$_s$} & \textbf{MBPP$_s$} & \textbf{GSM8K$_s$} & \textbf{RouterB.$_s$} \\
            \toprule
            
            \multirow{3}{*}{\textbf{Qwen 3B}} 
            & Base & 79.3 & 66.0 & 65.5 & 68.6 & 83.5 & 71.6 \\
            & Gem  & 84.1 & 67.1 & 67.8 & 73.3 & 89.0 & 73.2 \\
            & \textbf{\textcolor{ForestGreen}{Gain}} & \textbf{\textcolor{ForestGreen}{+4.8}} & \textbf{\textcolor{ForestGreen}{+1.1}} & \textbf{\textcolor{ForestGreen}{+2.3}} & \textbf{\textcolor{ForestGreen}{+4.7}} & \textbf{\textcolor{ForestGreen}{+5.5}} & \textbf{\textcolor{ForestGreen}{+1.6}} \\
            \midrule
            
            \multirow{3}{*}{\textbf{Qwen 7B}} 
            & Base & 90.0 & 67.1 & 71.3 & 80.9 & 89.7 & 78.3 \\
            & Gem  & 91.1 & 72.5 & 73.9 & 82.1 & 92.1 & 79.1 \\
            & \textbf{\textcolor{ForestGreen}{Gain}} & \textbf{\textcolor{ForestGreen}{+1.1}} & \textbf{\textcolor{ForestGreen}{+5.4}} & \textbf{\textcolor{ForestGreen}{+2.6}} & \textbf{\textcolor{ForestGreen}{+1.2}} & \textbf{\textcolor{ForestGreen}{+2.4}} & \textbf{\textcolor{ForestGreen}{+0.8}} \\
            \midrule

            \multirow{3}{*}{\textbf{Mistral 7B}} 
            & Base & 74.6 & 56.4 & 56.6 & 47.0 & 40.6 & 55.6 \\
            & Gem  & 81.8 & 66.9 & 66.2 & 58.8 & 80.7 & 69.6 \\
            & \textbf{\textcolor{ForestGreen}{Gain}} & \textbf{\textcolor{ForestGreen}{+7.2}} & \textbf{\textcolor{ForestGreen}{+10.5}} & \textbf{\textcolor{ForestGreen}{+9.6}} & \textbf{\textcolor{ForestGreen}{+11.8}} & \textbf{\textcolor{ForestGreen}{+40.1}} & \textbf{\textcolor{ForestGreen}{+14.0}} \\
            \midrule

            \multirow{3}{*}{\textbf{Llama 8B}} 
            & Base & 80.9 & 59.7 & 64.1 & 65.6 & 83.2 & 71.3 \\
            & Gem  & 86.0 & 67.1 & 71.6 & 68.8 & 96.0 & 74.7 \\
            & \textbf{\textcolor{ForestGreen}{Gain}} & \textbf{\textcolor{ForestGreen}{+5.1}} & \textbf{\textcolor{ForestGreen}{+7.4}} & \textbf{\textcolor{ForestGreen}{+7.5}} & \textbf{\textcolor{ForestGreen}{+3.2}} & \textbf{\textcolor{ForestGreen}{+12.8}} & \textbf{\textcolor{ForestGreen}{+3.4}} \\
            \bottomrule
        \end{tabular}
    }
    \label{tab:hidden_gems}
\end{table}

\begin{figure*}[t]
    \centering

    \begin{subfigure}{0.29\linewidth}
        \centering
        \includegraphics[width=\linewidth]{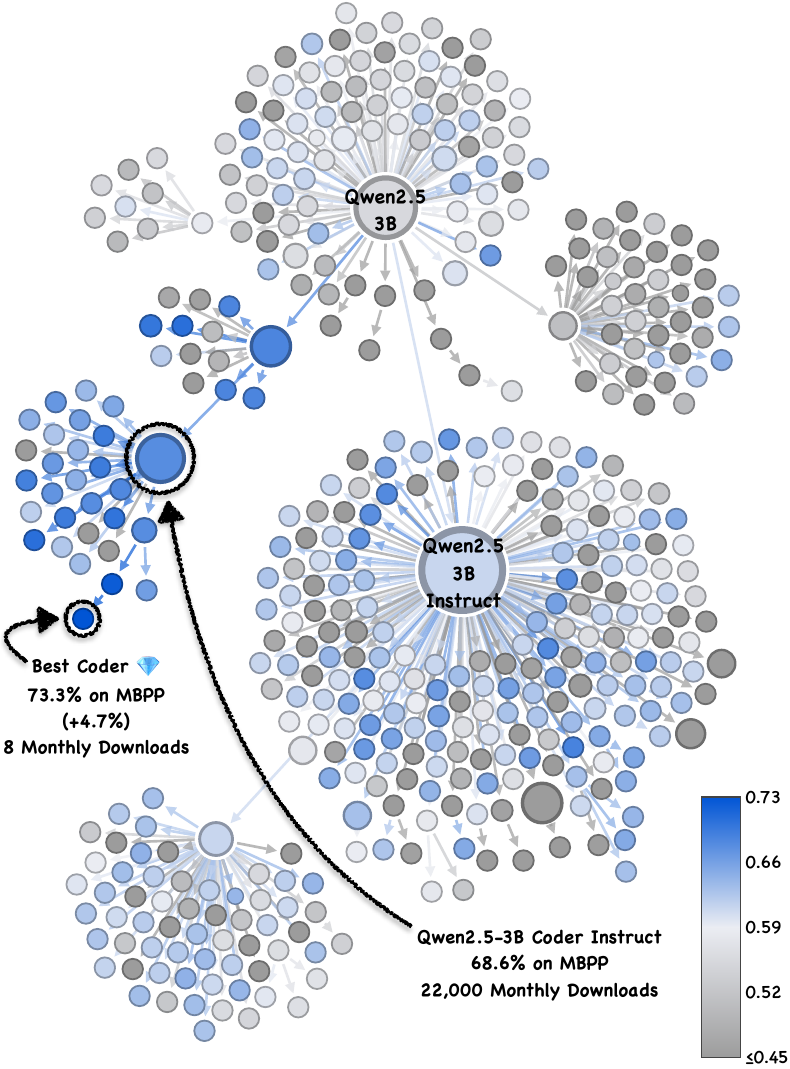}
        \caption{\textbf{\textit{Qwen-3B Coding Acc.}}}
    \end{subfigure}\hfill
    \begin{subfigure}{0.29\linewidth}
        \centering
        \includegraphics[width=\linewidth]{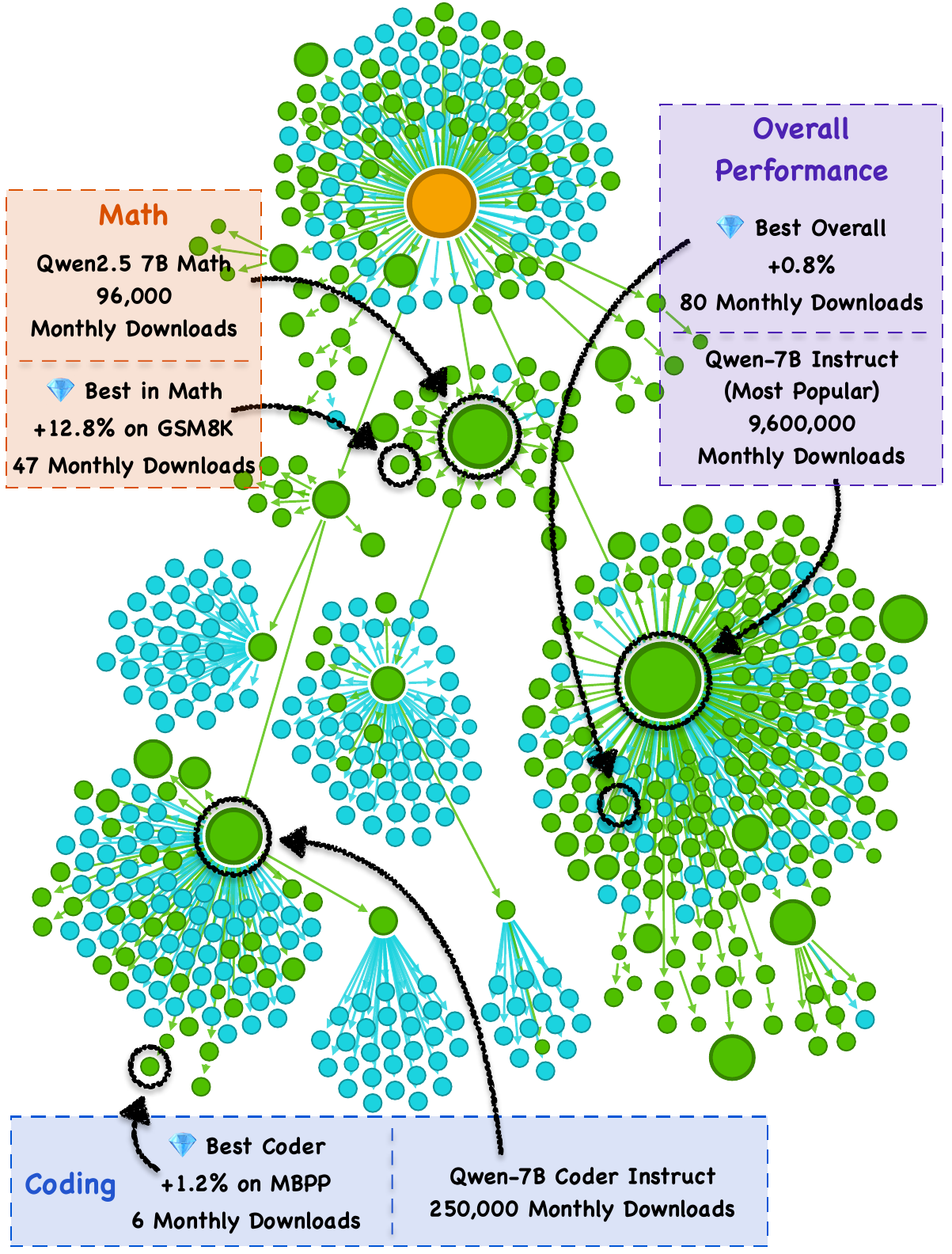}
        \caption{\textbf{\textit{Qwen 7B Hidden Gems.}}}
    \end{subfigure}\hfill
    \begin{subfigure}{0.29\linewidth}
        \centering
        \includegraphics[width=\linewidth]{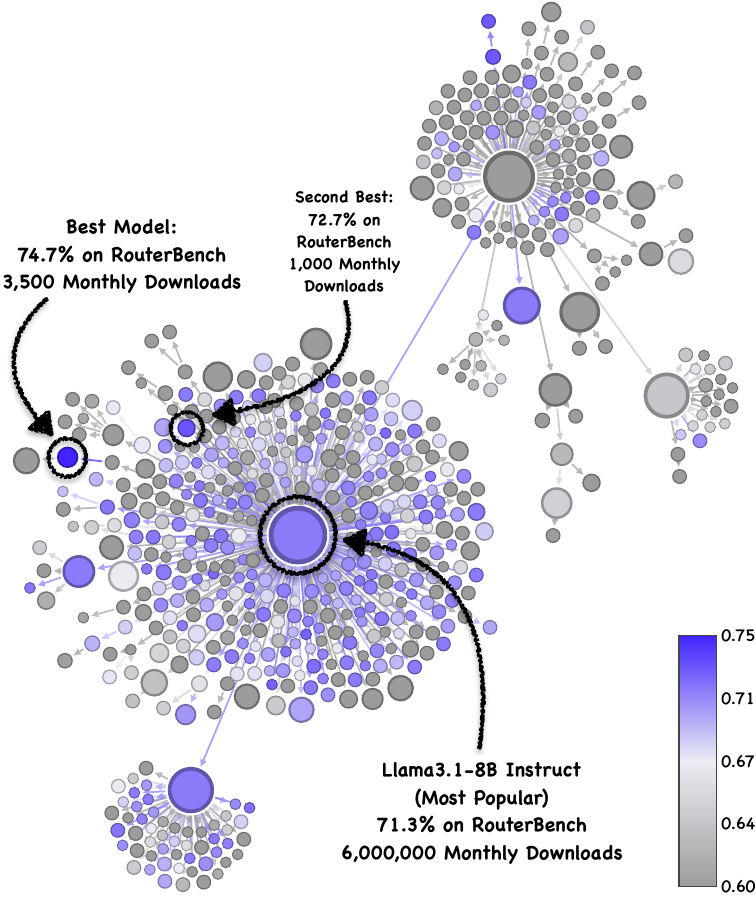}
        \caption{\textbf{\textit{Llama 3.1 8B RouterB.$_s$ Acc.}}}
    \end{subfigure}

    \caption{\textbf{\textit{Visualizing Hidden Gems in Model Trees.}} Nodes represent models, with size corresponding to monthly downloads in log-scale. (a) Qwen-3B tree colored by MBPP$_s$ coding performance; the best coder (circled) remains unpopular despite significantly outperforming the base instruct model. (b) Qwen-7B tree highlighting gems in math, coding and overall (RouterBench$_s$) performance. (c) Llama-3.1-8B tree colored by overall performance; the best-performing model has orders of magnitude fewer downloads than the official Llama3.1 8B Instruct version. 
    }
    \label{fig:atlas}
\end{figure*}

\section{Do Hidden Gems Exist?}
\label{sec:hidden_gems}

\noindent \textbf{The Centralization of Usage.} 
Public model repositories are expanding rapidly, with Hugging Face alone hosting over two million models. Despite this abundance, usage remains heavily centralized: As shown in Fig.~\ref{fig:usability}, a mere $0.0015\%$ of models account for $95\%$ of all downloads while the vast majority of the ecosystem is inactive. This disparity suggests two competing hypotheses. The \textit{Efficient Discovery} hypothesis posits that usage concentration is causal: users successfully converge to the best models. Alternatively, the \textit{Information Asymmetry} hypothesis argues that the best models remain \q{hidden} in the long tail, and that users have no real choice but using them due to a lack of reliable signals. Concretely, as evaluating millions of models on each task is infeasible, the default option becomes well-tested but sub-optimal checkpoints. 

\noindent \textbf{Defining Hidden Gems.} 
We propose a test to decide between these hypotheses. If we can identify unpopular models that strictly outperform the popular choice, we can reject the efficient discovery hypothesis. We term such models \textit{Hidden Gems}.

Formally, let $\mathcal{T} = \{m_1, \dots, m_K\}$ be a set of models. Let $r_i$ denote the performance of model $m_i$ on dataset $D$. We define the \textit{Popular Consensus} group, $\mathcal{P} \subset \mathcal{T}$, as the top $1\%$ of models by download count, and the \textit{Elite} group, $\mathcal{E}_D \subset \mathcal{T}$, as the top $1\%$ by performance. A model $m_i$ is a Hidden Gem if it satisfies three criteria: i) \emph{Obscurity:} $m_i \notin \mathcal{P}$ (it is unpopular); ii) \emph{Excellence:} $m_i \in \mathcal{E}_D$ (it is top-tier); and iii) \emph{Dominance:} $r_i > \max_{m_j \in \mathcal{P}} r_j$. 
Condition (iii) is crucial: Gems must strictly outperform the best popular models to justify the search.

\noindent \textbf{Popularity $\neq$ Performance.} 
We search for hidden gems by evaluating over $2,000$ models. To ensure identical inference costs and leave capability as the sole differentiator, we compare fine-tunes only within the same Model Trees \citep{horwitz2025unsupervised,Horwitz_2025_CVPR}, i.e., sets of fine-tunes derived from a shared ancestor. We test four major trees: Qwen2.5 (3B \& 7B) \citep{Yang2024Qwen25TR}, Mistral-7B \citep{Jiang2023Mistral7}, and Llama3.1-8B \citep{dubey2024llama}.
To assess general quality, we use RouterBench \citep{hu2024routerbench}, which aggregates diverse benchmarks including MBPP, Winograd, MMLU, ARC-Challenge, and GSM8K \citep{sakaguchi2021winogrande,clark2018think,hendrycks2020measuring,cobbe2021training,austin2021program}. Due to computational constraints, we evaluate on a subsampled set of $2,500$ randomly chosen queries, denoted by a subscript $s$, e.g., RouterBench$_s$ (see App.~\ref{app:implementation_details}). Unsurprisingly, usage is dominated by the official base releases; e.g., in the Qwen-3B tree, the official base and instruct variants alone capture over $80\%$ of all downloads. 

Tab.~\ref{tab:hidden_gems} compares the performance of the popular consensus against the best-performing gems. Across all trees and tasks, we consistently uncover models that are much better than their popular predecessors. 
A striking example is the Qwen-3B tree, where a math-oriented fine-tune boosts GSM8K$_s$ accuracy from $83.5\%$ up to $89.0\%$, nearing the performance of the best Qwen-7B base version, with less than half the parameters. 
Crucially, we reveal that hidden gems are not limited to niche tasks; every tree also includes unpopular models that are better \textit{generalists} than the popular base versions.

\noindent \textbf{The Failure of Heuristics.} 
Why are these models missed? We found that over $90\%$ of the identified gems lacked performance documentation relevant to their specific strengths, leaving users with no signal to identify them (see App.~\ref{app:documentation}). We then map the locations of these gems in Fig.~\ref{fig:atlas}. The best-performing models do not cluster near the root or along predictable trajectories. This implies that simple search heuristics based on popularity or graph centrality are likely to fail. These findings refute the efficient discovery hypothesis and motivate active search methods as we propose next.

\begin{table*}[t]
    \caption{\textbf{\textit{Model Discovery Results.}} We evaluate the top retrieval of each method. For each query budget, we report the mean rank and accuracy of the retrieved models for each model tree. We report the mean out of $100$ repetitions.}
    \centering

    {
    \tiny
    \setlength{\tabcolsep}{8pt}
    \renewcommand{\arraystretch}{1.05}
    \resizebox{\linewidth}{!}{
    \begin{tabular}{clccccccccc}
        \multirow{2}{*}{\shortstack{\textbf{Queries}\\\textbf{Per Model}}} & \multirow{2}{*}{\textbf{Method}} & \multicolumn{2}{c}{\textbf{Qwen-3B}} & \multicolumn{2}{c}{\textbf{Qwen-7B}} & \multicolumn{2}{c}{\textbf{Mistral-7B}} & \multicolumn{2}{c}{\textbf{Llama-8B}} \\
        \cmidrule(lr){3-4} \cmidrule(lr){5-6} \cmidrule(lr){7-8} \cmidrule(lr){9-10}
        &  & Rank $\downarrow$ & Acc. $\uparrow$ & Rank $\downarrow$ & Acc. $\uparrow$ & Rank $\downarrow$ & Acc. $\uparrow$ & Rank $\downarrow$ & Acc. $\uparrow$ \\
        \toprule
        \multirow{2}{*}{-} 
            & Random Selection 
                & 233.8 & 0.588
                & 318.4 & 0.599
                & 144.6 & 0.431
                & 270.8 & 0.601 \\
        & Best Base 
                & 56.5 & 0.716 
                & 30.0 & 0.783 
                & 62.0 & 0.556 
                & 41.0 & 0.713 \\
        \midrule

        \multirow{9}{*}{10} 
            & Uniform 
                & 166.3 & 0.671
                & 173.8 & 0.724
                & 29.5 & 0.656
                & 222.0 & 0.665 \\
            & UCB 
                & 88.0 & 0.708
                & 83.9 & 0.770
                & 14.1  & 0.684
                & 110.2 & 0.702 \\
            & UCB-StdDev. 
                & 81.8 & 0.710
                & 75.2 & 0.772
                & 11.8 & 0.686
                & 96.0 & 0.707 \\
            & TTTS 
                & 92.6 & 0.708
                & 75.7 & 0.772
                & 14.0 & 0.685
                & 103.6 & 0.702 \\
            & Successive Rejects 
                & 94.4 & 0.706
                & 128.3 & 0.752
                & 25.6 & 0.662
                & 171.5 & 0.683 \\
            & BayesElim 
                & 56.4 & 0.717
                & 58.9 & 0.778
                & 13.5 & 0.685
                & 79.9 & 0.708 \\
            & UCB-E 
                & 78.4 & 0.710
                & 83.7 & 0.769
                & 13.6 & 0.684
                & 106.5 & 0.703 \\
            & Sequential Halving 
                & 69.3 & 0.714
                & 62.9 & 0.777
                & 15.2 & 0.684
                & 75.2 & 0.710 \\
            & Ours 
                & \textbf{11.3} & \textbf{0.726}
                & \textbf{15.8} & \textbf{0.786}
                & \textbf{2.6} & \textbf{0.694}
                & \textbf{15.4} & \textbf{0.725} \\
        \midrule

        \multirow{9}{*}{50} 
            & Uniform 
                & 91.2 & 0.707
                & 81.9 & 0.770
                & 17.4 & 0.683
                & 114.9 & 0.705 \\
            & UCB 
                & 41.2 & 0.719
                & 39.8 & 0.782
                & 8.9  & 0.689
                & 55.8 & 0.718 \\
            & UCB-StdDev. 
                & 34.7 & 0.721
                & 28.9 & 0.784
                & 4.0  & 0.693
                & 37.1 & 0.722 \\
            & TTTS 
                & 53.1 & 0.717
                & 51.7 & 0.780
                & 10.8 & 0.687
                & 70.6 & 0.712 \\
            & Successive Rejects 
                & 83.1 & 0.710
                & 78.4 & 0.773
                & 16.5 & 0.682
                & 92.0 & 0.707 \\
            & BayesElim 
                & 30.0 & 0.721
                & 31.0 & 0.784
                & 6.7 & 0.691
                & 34.4 & 0.721 \\
            & UCB-E 
                & 37.4 & 0.720
                & 33.1 & 0.783
                & 4.3 & 0.693
                & 43.4 & 0.720 \\
            & Sequential Halving 
                & 41.0 & 0.719
                & 29.6 & 0.784
                & 7.7 & 0.690
                & 29.9 & 0.720 \\
            & Ours 
                & \textbf{3.5}  & \textbf{0.729}
                & \textbf{3.6}  & \textbf{0.790}
                & \textbf{1.6}  & \textbf{0.695}
                & \textbf{3.0}  & \textbf{0.736} \\
        \midrule

        & Best Available
            & 1.0 & 0.732 & 1.0 & 0.791 & 1.0 & 0.696 & 1.0 & 0.747 \\
        \bottomrule
    \end{tabular}
    }
    }
    \label{tab:model_discovery}
\end{table*}

\section{Efficient Model Discovery}

\noindent \textbf{Problem Formulation.} Evaluating millions of models may require billions of queries which is infeasible. We therefore formulate practical model discovery as a budget-constrained problem. Given a model tree $\mathcal{T} = \{m_1, \dots, m_K\}$ and dataset $\mathcal{D}$, our goal is to identify the best-performing model $m^*$ under a global query budget $B$.

\noindent \textbf{Best Arm Identification (BAI).}
We frame this as a Fixed-Budget Best-Arm Identification \citep{audibert2010best} problem: the \emph{Arms} are candidate models in $\mathcal{T}$; an \emph{Action} is a single query evaluation; and \emph{Rewards} are binary ($+1$ for correct, else $0$). Unlike standard MAB, where an agent maximizes total reward during learning, our goal is pure exploration. We aim to minimize the \textit{simple regret} (gap between the chosen model and the best one) after the fixed budget $B$ is exhausted.

\noindent \textbf{Sequential Halving (SH) \citep{karnin2013almost}.} We adopt SH as our base algorithm. SH operates in rounds $s=1 \dots S$. First, it uniformly allocates a small budget ($5$-$25$ queries) to all models. At the end of each round, models are ranked by empirical accuracy; the bottom $50\%$ are eliminated, and the survivors are tested with more queries. This repeats until a single model remains. While SH is theoretically sound, standard implementations are sample-inefficient in our domain. We therefore introduce two domain-specific modifications. We illustrate our proposed approach in Fig.~\ref{fig:mobe_search}.

\noindent \textbf{Variance Reduction via Correlated Sampling.}
A standard SH implementation samples queries independently for each model. However, benchmarks contain a mix of trivial and even unsolvable problems. If Model A is tested with \q{easy} queries and Model B with \q{hard} ones, the ranking can become noisy. To mitigate this, we employ \textit{Correlated Sampling}. In every round, we enforce that all surviving models are tested with the exact same queries, thus minimizing the variance of the difference estimator.

\noindent \textbf{Aggressive Elimination Schedule.}
We present the distribution of model quality in Fig.~\ref{fig:distributions}, where we notice that the vast majority of uploads are low-quality or broken. These under-performers can be detected with very few queries. Standard SH, which eliminates only $50\%$ of candidates per round, wastes budget refining estimates for obviously poor models. We therefore modify the schedule to perform aggressive elimination already in the first round. Specifically, we reduce the candidate pool to a fixed size of $100$ models immediately. This "fail-fast" strategy reallocates a larger chunk of the compute budget to distinguishing between the elite candidates in later rounds. For more details see App.~\ref{app:implementation_details}.
App.~\ref{app:ablations} confirms our modified scheduler is highly important, significantly boosting retrieval ranks and accuracies through improved query allocation.

\section{Experiments}

\noindent \textbf{Experimental Setting.} We evaluate our method and baselines on searching for the optimal model $m^*$. We use four distinct model pools derived from the Qwen-3B, Qwen-7B, Mistral-7B, and Llama3-8B trees. We define the total query budget as $B = N \times K$, where $K$ is the number of models and $N \in \{10, 50\}$ is the average queries per model (for more see App.~\ref{app:additional_results}.). We repeat each experiment $100$ times and report the mean rank and top-1 accuracy. 

\noindent \textbf{Baselines.} We compare against $8$ established algorithms. \textit{Uniform} allocates budget equally. \textit{UCB} \citep{auer2002finite}, \textit{UCB-E} \citep{audibert2010best}, and \textit{UCB-StdDev} greedily sample based on upper confidence bounds. \textit{Successive Rejects} \citep{audibert2010best} and \textit{Sequential Halving} \citep{karnin2013almost} iteratively eliminate the worst models. We also include \textit{TTTS} \citep{russo2016simple}, which uses top-two Thompson sampling, and \textit{Bayesian Elimination} \citep{atsidakou2022bayesian}, which prunes based on posterior probabilities.

\noindent \textbf{Results.} 
Table~\ref{tab:model_discovery} summarizes the performance for $10, 50$ queries per model budgets (for more see App.~\ref{app:additional_results}). In the low-budget regime ($N=10$), standard baselines struggle where in the Qwen and Llama trees, most methods fail to even outperform the popular base versions. The exception is the Mistral-7B tree, where many models surpass the base (see App.~\ref{app:distributions}). In contrast, our method consistently identifies \q{Hidden Gems} that significantly outperform the popular consensus across all trees.  

In the mid-budget regime ($N=50$), elimination-based baselines begin to improve, occasionally beating the base models, but often finding models significantly weaker than the best available. Our approach, uses correlated sampling and aggressive pruning and consistently converges to a top-$3$ model with much better accuracy. See App.~\ref{app:ablations} for ablations confirming both design choices.

\section{Conclusion} We demonstrated that public repositories contain \q{Hidden Gems}, unpopular models that strictly outperform foundation checkpoints. To address the computational intractability of finding them, we proposed an accelerated Sequential Halving algorithm. Our method consistently identifies elite models with only $50$ queries per candidate, making model selection practically achievable.

\clearpage

\bibliography{custom}

\clearpage

\appendix

\section{Limitations}

\paragraph{Evaluation Resources.} While our approach expedites the search by over $50\times$, it still requires evaluating all models on a small number of queries. An alternative direction is weight-space learning, which aims to learn semantic representations of model weights, or subsets of weights, and retrieve models directly in weight space \citep{kahana2025can,Horwitz_2025_CVPR,lu2023content}. Although weight-space learning has the potential to further accelerate discovery, it is currently limited to small-scale networks and benchmarks.

\paragraph{Other Tasks.} While we found hidden gems in math, coding, question answering and general performance, we did not evaluate all possible tasks. Finding gems for new task would require to re-evaluate the models on this task.

\begin{figure}
    \centering
    \includegraphics[width=\linewidth]{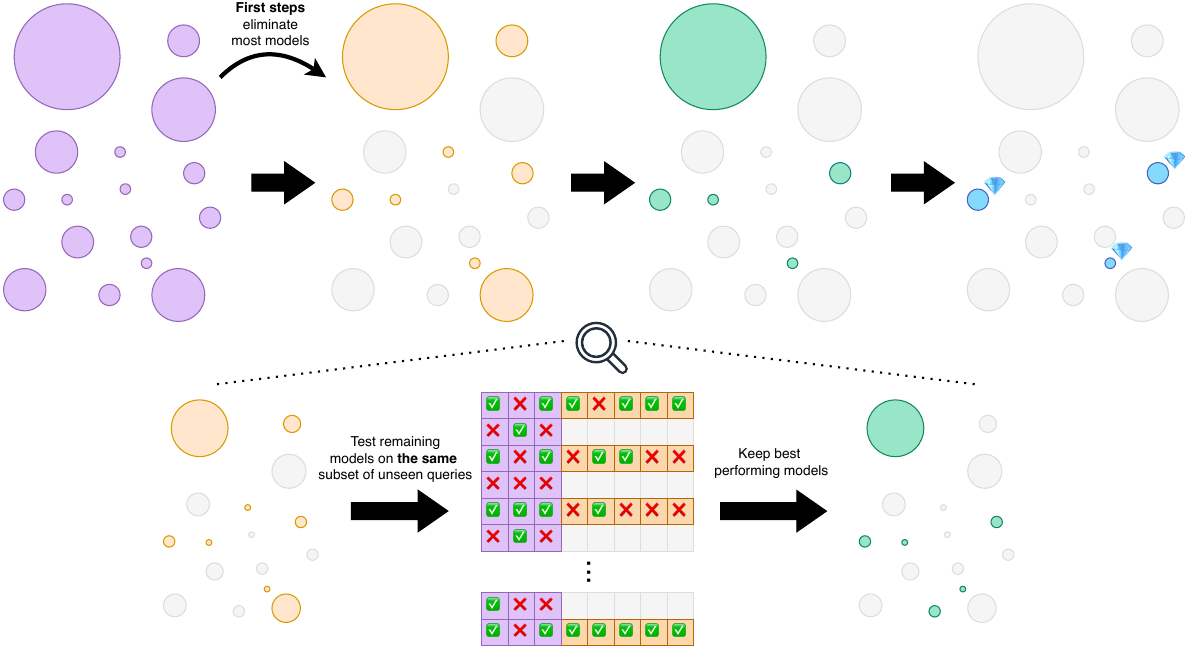}
    \caption{Our proposed Model Search Algorithm}
    \label{fig:mobe_search}
\end{figure}

\section{Related Works and Model Populations}
 Selecting an appropriate model for a given use case has long been a central challenge in machine learning \citep{zamir2018taskonomy,lu2023content,zhang2023model,luo2024stylus,tran2019transferability,nguyen2020leep,choshen-etal-2023-start,muqeeth2024learning}. The recent explosive growth in the number of publicly available models has further intensified this challenge, catalyzing a growing body of work on model populations. One prominent direction is \textit{Weight-Space Learning}, which aims to train neural networks that treat other neural networks not only as functions, but as data points \citep{schurholt2024neural,schurholt2021self,schurholt2022hyper,horwitz2025we,pal2024model}. The objective is to predict a wide range of model attributes \citep{salama2025dataset}, including performance \citep{kahana2025deep,nws}, functionality \citep{Horwitz_2025_CVPR,falk2025learning,gueta2023knowledge,schmidhuber}, lineage \citep{horwitz2025unsupervised,zhu2025independence,neural_phylogeny,phylolm,neural_lineage}, and training trajectories \citep{horwitz2024recovering,kuditipudi2025blackbox}, directly from model weights, without access to in distribution data. More broadly, weight space learning seeks to learn compact, structured representations of neural networks that enable search \citep{kahana2025can}, comparison, and discovery directly in weight space. This is challenging due to the extreme dimensionality of model parameters and the presence of many parameter space symmetries \citep{hecht1990algebraic,zhao2025symmetry}, such as permutation invariances \citep{lim2024empirical,shamsian2024improved,lim2023graph,lol,navon2023equivariant_alignment,navon2023equivariant,kofinas2024graph,neural_functional_transformers,zhou2024universal,zhou2024permutation}.

Beyond predicting attributes of individual models, recent works study model populations as structured objects in their own right \citep{horwitz2025we,wang2025ml}. This line of research asks how populations are organized, what insights can be derived from their global structure, how to identify gaps or communities within the population \citep{horwitz2025we}, and how AI model supply chains \citep{hopkins2025ai,jia2025a,EcosystemGraphs} shape the distribution of available models \citep{yang2024navigating,osborne2024ai,gao2023origin,longpre2025economies}.

Model populations have also been leveraged as a resource for improving existing models. A notable example is model merging \citep{ties_merging,stoica2024model,avrahami2022gan,shah2023ziplora,wortsman2022model}, where the weights of two or more models are combined to produce a model with improved or complementary capabilities. More broadly, population level approaches have been explored for collaborative optimization \citep{feng2024model,feng2025data,don-yehiya-etal-2023-cold}, collaborative generation, weight generation \citep{weights2weights,peebles2022learning,ha2016hypernetworks,erkocc2023hyperdiffusion}, and interpretability \citep{weights2weights,dravid2023rosetta,bar2025beyond,barshalom2025tokenprobabilitieslearnablefast,clip_dissect}. In contrast, we show that even without applying such population level methods, existing model populations already contain highly valuable models that can substantially outperform commonly used models.

\section{Implementation Details} 
\label{app:implementation_details}

\paragraph{Model Selection.} 
We select models from $4$ highly popular model trees created from $4$ popular base models: Qwen2.5-3B, Qwen2.5-7B, Mistral-7B and Llama3.1-8B. As these trees are very large we cannot evaluate them all. Therefore, for each tree we select $400$ downstream fine-tunings of the base model (including itself) and $400$ adapters (if existing). To ensure we dont miss a high-performing popular version, we select the $300$ most popular models of the tree first and then an additional $100$ random ones for both the fine-tunings and adapters. However, many models fail to download or run due to various mismatches, such as missing weight tensors or hugging-face library non-support for older models. We therefore are left with much less models in each tree. We detail the number of models in each tree in Tab.~\ref{tab:model_trees}. 

\begin{table}[h!]
    \centering
    \begin{tabular}{ccc}
         \textbf{Tree} & \textbf{\#Fine-tunes} & \textbf{\#Adapters} \\
         \toprule
         Qwen-2.5 (3B) & 331 & 148 \\
         Mistral (7B) & 240 & 40 \\
         Qwen-2.5 (7B) & 317 & 333 \\
         Llama-3.1 (8B) & 320 & 297 \\
    \end{tabular}
    \caption{\textbf{\textit{Number Evaluated Models from Each Tree.}}}
    \label{tab:model_trees}
\end{table}

\begin{table}[ht]
    \centering
    \caption{\textbf{\textit{Our custom scheduler.}} Queries per-model for different rounds ($s_1 \dots s_5$) across different budgets.}
    \label{tab:scheduler}
    \begin{tabular}{@{}cccccc@{}}
        \toprule
        \multirow{2}{*}{\textbf{Budget}} & \multicolumn{5}{c}{\textbf{Queries-per-model}} \\ \cmidrule(lr){2-6} 
         & \textbf{$s_1$} & \textbf{$s_2$} & \textbf{$s_3$} & \textbf{$s_4$} & \textbf{$s_5$} \\ \midrule
        10  & 6   & 16  & 30  & 60  & 120  \\
        25  & 15  & 40  & 75  & 150 & 300  \\
        50  & 30  & 75  & 150 & 300 & 600  \\
        100 & 60  & 150 & 300 & 600 & 1200 \\
        200 & 120 & 300 & 600 & 1200 & 2400 \\ \bottomrule
    \end{tabular}
\end{table}

\paragraph{Query Selection.} To facilitate the evaluation of over $2,000$ models, we subsample the RouterBench evaluation suite to a total of $2,500$ queries and allocate the budget uniformly across the tasks (ARC-C, Winogrande, MMLU, MBPP, and GSM8K). Queries within each task are selected via random sampling to ensure an unbiased representation of the original benchmarks.

\paragraph{Model Evaluation.} To ensure a fair comparison, we wish to evaluate all models under consistent conditions. However, we observed that some models might benefit from using their system prompt and some would not. Moreover, there could be inconsistencies within the same model across different tasks. To overcome these inconsistencies, we evaluated each model both with and without its system prompt and, given a task, took the version with the best accuracy for it. 

For reproducibility, we evaluate models by greedy decoding (\texttt{do\_sample=False}). Lastly, we set a maximum response length of \texttt{max\_length=50} tokens at multiple-choice and short answer queries (ARC-Challenge, MMLU and Winogrande) which typically require only a single response token. In coding (MBPP) and math (GSM8K) queries, which may require longer answers or some thinking, we set a limit of \texttt{max\_length=512} tokens.

\begin{table*}[t!]
    \caption{\textbf{\textit{Ablation Studies.}} We ablate both parts of our approach. We report the mean rank and accuracy of the retrieved models for each model tree. Each experiment is repeated $100$ times.}
    \vspace{0.2cm}
    \centering

    {
    \small
    \setlength{\tabcolsep}{8pt}
    \renewcommand{\arraystretch}{1.05}
        \resizebox{\linewidth}{!}{
    \begin{tabular}{ccccccccccc}
        \multirow{2}{*}{\shortstack{\textbf{Queries}\\\textbf{Per Model}}} & \multirow{2}{*}{\textbf{Method}} & \multicolumn{2}{c}{\textbf{Qwen-3B}} & \multicolumn{2}{c}{\textbf{Qwen-7B}} & \multicolumn{2}{c}{\textbf{Mistral-7B}} & \multicolumn{2}{c}{\textbf{Llama-8B}} \\
        \cmidrule(lr){3-4} \cmidrule(lr){5-6} \cmidrule(lr){7-8} \cmidrule(lr){9-10}
        &  & Rank $\downarrow$ & Acc. $\uparrow$ & Rank $\downarrow$ & Acc. $\uparrow$ & Rank $\downarrow$ & Acc. $\uparrow$ & Rank $\downarrow$ & Acc. $\uparrow$ \\
        \toprule
        \multirow{2}{*}{-} 
            & Random Selection 
                & 233.8 & 0.588
                & 318.4 & 0.599
                & 144.6 & 0.431
                & 270.8 & 0.601 \\
        & Best Base 
                & 56.5 & 0.716 
                & 30.0 & 0.783 
                & 62.0 & 0.556 
                & 41.0 & 0.713 \\
        \midrule

        \multirow{4}{*}{10} 
            & Sequential Halving 
                & 69.3 & 0.714
                & 62.9 & 0.777
                & 15.2 & 0.684
                & 75.2 & 0.710 \\
            & Ours w. SH Scheduler 
                & 51.9 & 0.717
                & 45.4 & 0.780
                & 12.7 & 0.684
                & 56.2 & 0.718 \\
            & Ours w/o Corr. Sampling  
                & 12.6 & 0.725
                & \textbf{10.4} & \textbf{0.788}
                & 2.8 & \textbf{0.694}
                & \textbf{11.1} & \textbf{0.725} \\
            & Ours 
                & \textbf{11.3} & \textbf{0.726}
                & 15.8 & 0.786
                & \textbf{2.6} & \textbf{0.694}
                & 15.4 & \textbf{0.725} \\
        \midrule

        \multirow{4}{*}{50} 
            & Sequential Halving 
                & 41.0 & 0.719
                & 29.6 & 0.784
                & 7.7 & 0.690
                & 29.9 & 0.720 \\
            & Ours w. SH Scheduler 
                & 10.6 & 0.726
                & 13.5 & 0.787
                & 5.3 & 0.692
                & 11.2 & 0.731 \\
            & Ours w/o Corr. Sampling 
                & 5.4 & 0.728
                & 5.5 & 0.789
                & 2.0 & \textbf{0.695}
                & 6.5 & 0.729 \\
            & Ours 
                & \textbf{3.5}  & \textbf{0.729}
                & \textbf{3.6}  & \textbf{0.790}
                & \textbf{1.6}  & \textbf{0.695}
                & \textbf{3.0}  & \textbf{0.736} \\
        \midrule

        & Best Available
            & 1.0 & 0.732 & 1.0 & 0.791 & 1.0 & 0.696 & 1.0 & 0.747 \\
        \bottomrule
    \end{tabular}
    }
    }
    \label{tab:ablations}
\end{table*}

\paragraph{Scheduler.} Driven by models' skewed accuracy distributions(see App.~\ref{app:distributions}), we design a custom scheduler for efficient model discovery. First, as the vast majority of models are significantly worse than the best available option, we can eliminate them quickly. Therefore, we make an aggressive first elimination round of keeping a fixed number of just $100$ models, which translates to approximately $20\%$ of the model tree on average. However, it is critical we do not eliminate high-performing models in this round. Therefore, we also modify the query allocation schedule between rounds. Specifically, we allocate $60\%$ of the budget to the first step, making it the most resource intensive. As we are working with highly constraint budgets of up to just $10$ queries per model, this translates to as few as $6$ evaluations. We then double the amount of queries per-model at each round. Using this scheduler, we avoid early eliminations of good models, while leaving a substantial amount of the budget to the remaining ones as their number quickly decreases. We detail our exact scheduler for different budgets at Tab.~\ref{tab:scheduler}. Finally,
Tab.~\ref{tab:ablations} and Tab.~\ref{tab:additional_results} show this scheduler is highly effective, and greatly improves the accuracy and rank of the detected models.

\section{Ablation Studies}
\label{app:ablations}
We ablate the main two parts of our method: (i) correlated sampling and (ii) elimination schedule. We present the results in Tab.~\ref{tab:ablations}. For our modified scheduler, the results are clear: our scheduler can retrieve significantly better models under the same budget, focusing the search on the more \q{interesting} comparisons. Specifically, at a $10$ queries (per-model) budget we observe an retrieval rank decrease of over $30$ when averaged across trees using our custom schedule. However, this is less clear in the case of correlated sampling. In highly small budgets, we find both options to be comparable, as the evaluation is very noisy in the first place. However, this changes as the budget increases: in a larger budget of $50$ we can see that in all trees, correlated sampling achieves better results, allowing our approach to retrieve a top-$3$ available model.

\section{Additional Budgets Results}
\label{app:additional_results}
We evaluate our method's robustness across three additional query budgets: $25$, $100$, and $200$ queries per model. Results are presented in Tab.~\ref{tab:additional_results}. 
As expected, increasing the budget size leads to saturated performance, where the baseline approaches also begin to consistently identify gem models that are significantly better than the best popular base versions. Nevertheless, our approach remains superior, often matching or surpassing the baselines performance while requiring only half the query budget. For instance, our method's performance with a budget of $25$ queries per-model exceeds that of all baselines given a doubled budget of $50$. We observe a similar trend when comparing our $50$ query results against the baselines $100$ query results.
Furthermore, under equal budget constraints, our approach consistently identifies superior models with higher average accuracies and lower average ranks.

\begin{table}[h!]
    \caption{\textbf{\textit{Gem Documentation.}} We analyze which identified hidden gems include performance documentation. 
    \docYes~indicates relevant documentation, 
    \docNo~indicates no documentation, and 
    \docIrr~indicates documentation for irrelevant tasks only (e.g., multilingual scores for a math model).}
    \centering
    \resizebox{\linewidth}{!}{
        \begin{tabular}{ccccccc}
            \textbf{Tree} & \textbf{ARC-C$_s$} & \textbf{WinoG.$_s$} & \textbf{MMLU$_s$} & \textbf{MBPP$_s$} & \textbf{GSM8K$_s$} & \textbf{RouterB.$_s$} \\
            \toprule
            Qwen 3B 
            & \docIrr & \docNo & \docNo & \docYes & \docNo & \docNo \\
            \midrule
            Mistral 7B 
            & \docNo & \docNo & \docNo & \docNo & \docNo & \docNo \\
            \midrule
            Qwen 7B 
            & \docNo & \docNo & \docNo & \docNo & \docNo & \docNo \\
            \midrule
            Llama 8B 
            & \docNo & \docNo & \docIrr & \docNo & \docYes & \docIrr \\
            \bottomrule
        \end{tabular}
    }
    \label{tab:hidden_gems_doc}
\end{table}

\begin{table*}[t]
    \caption{\textbf{\textit{Extended Model Discovery Results.}} We evaluate the top retrievals of each method. For each budget (10, 25, 50, 100, 200), we report the rank and accuracy of the retrieved models for each model tree. Each experiment is repeated $100$ times, and mean results are reported.}
    \vspace{0.2cm}
    \centering
    {
    \small
    \setlength{\tabcolsep}{8pt}
    \renewcommand{\arraystretch}{1.05}
    
    \begin{tabular}{lcccccccccc}
        \textbf{Queries} &  & \multicolumn{2}{c}{\textbf{Qwen-3B}} & \multicolumn{2}{c}{\textbf{Qwen-7B}} & \multicolumn{2}{c}{\textbf{Mistral-7B}} & \multicolumn{2}{c}{\textbf{Llama-8B}} \\
        \cmidrule(lr){3-4} \cmidrule(lr){5-6} \cmidrule(lr){7-8} \cmidrule(lr){9-10}
        \textbf{Per Model} & \textbf{Method} & Rank $\downarrow$ & Acc. $\uparrow$ & Rank $\downarrow$ & Acc. $\uparrow$ & Rank $\downarrow$ & Acc. $\uparrow$ & Rank $\downarrow$ & Acc. $\uparrow$ \\
        \toprule
        \multirow{2}{*}{-} 
            & Random Selection 
                & 233.8 & 0.588
                & 318.4 & 0.599
                & 144.6 & 0.431
                & 270.8 & 0.601 \\
        & Best Base 
                & 56.5 & 0.716 
                & 30.0 & 0.783 
                & 62.0 & 0.556 
                & 41.0 & 0.713 \\
        \midrule

        \multirow{9}{*}{10} 
            & Uniform
                & 166.3 & 0.671
                & 173.8 & 0.724
                & 29.5 & 0.656
                & 222.0 & 0.665 \\
            & UCB 
                & 88.0 & 0.708
                & 83.9 & 0.770
                & 14.1  & 0.684
                & 110.2 & 0.702 \\
            & UCB-StdDev. 
                & 81.8 & 0.710
                & 75.2 & 0.772
                & 11.8 & 0.686
                & 96.0 & 0.707 \\
            & TTTS 
                & 92.6 & 0.708
                & 75.7 & 0.772
                & 14.0 & 0.685
                & 103.6 & 0.702 \\
            & Successive Rejects 
                & 94.4 & 0.706
                & 128.3 & 0.752
                & 25.6 & 0.662
                & 171.5 & 0.683 \\
            & BayesElim 
                & 56.4 & 0.717
                & 58.9 & 0.778
                & 13.5 & 0.685
                & 79.9 & 0.708 \\
            & UCB-E 
                & 78.4 & 0.710
                & 83.7 & 0.769
                & 13.6 & 0.684
                & 106.5 & 0.703 \\
            & Sequential Halving 
                & 69.3 & 0.714
                & 62.9 & 0.777
                & 15.2 & 0.684
                & 75.2 & 0.710 \\
            & Ours 
                & \textbf{11.3} & \textbf{0.726}
                & \textbf{15.8} & \textbf{0.786}
                & \textbf{2.6} & \textbf{0.694}
                & \textbf{15.4} & \textbf{0.725} \\
        \midrule

        \multirow{9}{*}{25} 
            & Uniform 
                & 115.6 & 0.697
                & 126.4 & 0.752
                & 19.4 & 0.679
                & 167.9 & 0.687 \\
            & UCB 
                & 63.6 & 0.715
                & 64.7 & 0.777
                & 12.4  & 0.686
                & 74.1 & 0.712 \\
            & UCB-StdDev. 
                & 54.9 & 0.717
                & 53.0 & 0.780
                & 8.8 & 0.689
                & 55.4 & 0.718 \\
            & TTTS 
                & 76.4 & 0.712
                & 62.5 & 0.777
                & 15.3 & 0.684
                & 96.2 & 0.705 \\
            & Successive Rejects 
                & 88.6 & 0.707
                & 76.7 & 0.773
                & 19.1 & 0.678
                & 122.3 & 0.699 \\
            & BayesElim 
                & 47.2 & 0.719
                & 43.9 & 0.782
                & 9.8 & 0.689
                & 56.3 & 0.715 \\
            & UCB-E 
                & 55.0 & 0.717
                & 58.4 & 0.777
                & 7.6 & 0.690
                & 63.7 & 0.715 \\
            & Sequential Halving 
                & 52.4 & 0.717
                & 35.7 & 0.783
                & 11.6 & 0.687
                & 50.9 & 0.717 \\
            & Ours 
                & \textbf{5.3}  & \textbf{0.728}
                & \textbf{12.3}  & \textbf{0.787}
                & \textbf{1.9}  & \textbf{0.695}
                & \textbf{8.0}  & \textbf{0.730} \\
        \midrule

        \multirow{9}{*}{50} 
            & Uniform 
                & 91.2 & 0.707
                & 81.9 & 0.770
                & 17.4 & 0.683
                & 114.9 & 0.705 \\
            & UCB 
                & 41.2 & 0.719
                & 39.8 & 0.782
                & 8.9  & 0.689
                & 55.8 & 0.718 \\
            & UCB-StdDev. 
                & 34.7 & 0.721
                & 28.9 & 0.784
                & 4.0  & 0.693
                & 37.1 & 0.722 \\
            & TTTS 
                & 53.1 & 0.717
                & 51.7 & 0.780
                & 10.8 & 0.687
                & 70.6 & 0.712 \\
            & Successive Rejects 
                & 83.1 & 0.710
                & 78.4 & 0.773
                & 16.5 & 0.682
                & 92.0 & 0.707 \\
            & BayesElim 
                & 30.0 & 0.721
                & 31.0 & 0.784
                & 6.7 & 0.691
                & 34.4 & 0.721 \\
            & UCB-E 
                & 37.4 & 0.720
                & 33.1 & 0.783
                & 4.3 & 0.693
                & 43.4 & 0.720 \\
            & Sequential Halving 
                & 41.0 & 0.719
                & 29.6 & 0.784
                & 7.7 & 0.690
                & 29.9 & 0.720 \\
            & Ours 
                & \textbf{3.5}  & \textbf{0.729}
                & \textbf{3.6}  & \textbf{0.790}
                & \textbf{1.6}  & \textbf{0.695}
                & \textbf{3.0}  & \textbf{0.736} \\
        \midrule

        \multirow{9}{*}{100} 
            & Uniform 
                & 63.2 & 0.714
                & 60.7 & 0.777
                & 14.5 & 0.685
                & 86.2 & 0.713 \\
            & UCB 
                & 30.6 & 0.721
                & 29.2 & 0.784
                & 3.5  & 0.694
                & 27.7 & 0.728 \\
            & UCB-StdDev. 
                & 16.8 & 0.725
                & 19.9 & 0.787
                & 2.3  & \textbf{0.695}
                & 13.2 & 0.734 \\
            & TTTS 
                & 47.2 & 0.719
                & 35.5 & 0.783
                & 5.6 & 0.692
                & 46.6 & 0.718 \\
            & Successive Rejects 
                & 69.3 & 0.714
                & 57.3 & 0.779
                & 13.8 & 0.685
                & 63.3 & 0.713 \\
            & BayesElim 
                & 10.9 & 0.726
                & 16.5 & 0.786
                & 3.4 & 0.694
                & 11.6 & 0.729 \\
            & UCB-E 
                & 12.4 & 0.726
                & 14.7 & 0.787
                & 2.8 & 0.694
                & 13.2 & 0.735 \\
            & Sequential Halving 
                & 17.9 & 0.724
                & 16.1 & 0.786
                & 3.9 & 0.693
                & 14.8 & 0.726 \\
            & Ours 
                & \textbf{2.1}  & \textbf{0.731}
                & \textbf{2.0}  & \textbf{0.791}
                & \textbf{1.4}  & \textbf{0.695}
                & \textbf{1.9}  & \textbf{0.740} \\
        \midrule

        \multirow{9}{*}{200} 
            & Uniform 
                & 30.5 & 0.722
                & 45.0 & 0.781
                & 10.6 & 0.688
                & 54.8 & 0.723 \\
            & UCB 
                & 20.0 & 0.726
                & 20.4 & 0.786
                & 1.2  & \textbf{0.696}
                & 8.4  & 0.741 \\
            & UCB-StdDev. 
                & 10.0 & 0.729 & 6.3 & 0.789 
                & 1.3 & 0.695
                & 1.8 & 0.743 \\
            & TTTS 
                & 23.4 & 0.723
                & 30.0 & 0.784
                & 1.2 & \textbf{0.696}
                & 16.1 & 0.734 \\
            & Successive Rejects 
                & 52.4 & 0.717
                & 52.0 & 0.780
                & 7.3 & 0.690
                & 52.2 & 0.716 \\
            & BayesElim 
                & 4.7 & 0.728
                & 5.5 & 0.789
                & 2.0 & 0.695
                & 4.0 & 0.738 \\
            & UCB-E 
                & 5.7 & 0.730
                & 4.2 & 0.790
                & 2.0 & 0.695
                & 1.9 & 0.743 \\
            & Sequential Halving 
                & 6.2 & 0.728
                & 7.3 & 0.788
                & 2.3 & 0.695
                & 4.3 & 0.739 \\
            & Ours 
                & \textbf{1.4} & \textbf{0.731}
                & \textbf{1.7} & \textbf{0.791}
                & \textbf{1.1} & \textbf{0.696}
                & \textbf{1.3} & \textbf{0.744} \\
        \midrule
            & Optimal 
                & 1.0 & 0.732 & 1.0 & 0.791 & 1.0 & 0.696 & 1.0 & 0.747 \\
        \bottomrule
    \end{tabular}
    }
    \label{tab:additional_results}
\end{table*}

\section{Model Accuracy Distributions}
\label{app:distributions}
We present the accuracy distributions across different models trees for math, coding and overall performances. For each tree and task we plot the cumulative histogram, and mark $2$ important lines: one at $-10\%$ from the best gem model, and another at $-5\%$. In all tasks and trees the vast majority of models are more than $10\%$ worse than the best model, and thus can be easily detected and ignored.

\begin{figure*}[t]
    \centering
    \begin{tabular}{ccc}

        \includegraphics[width=0.31\linewidth]{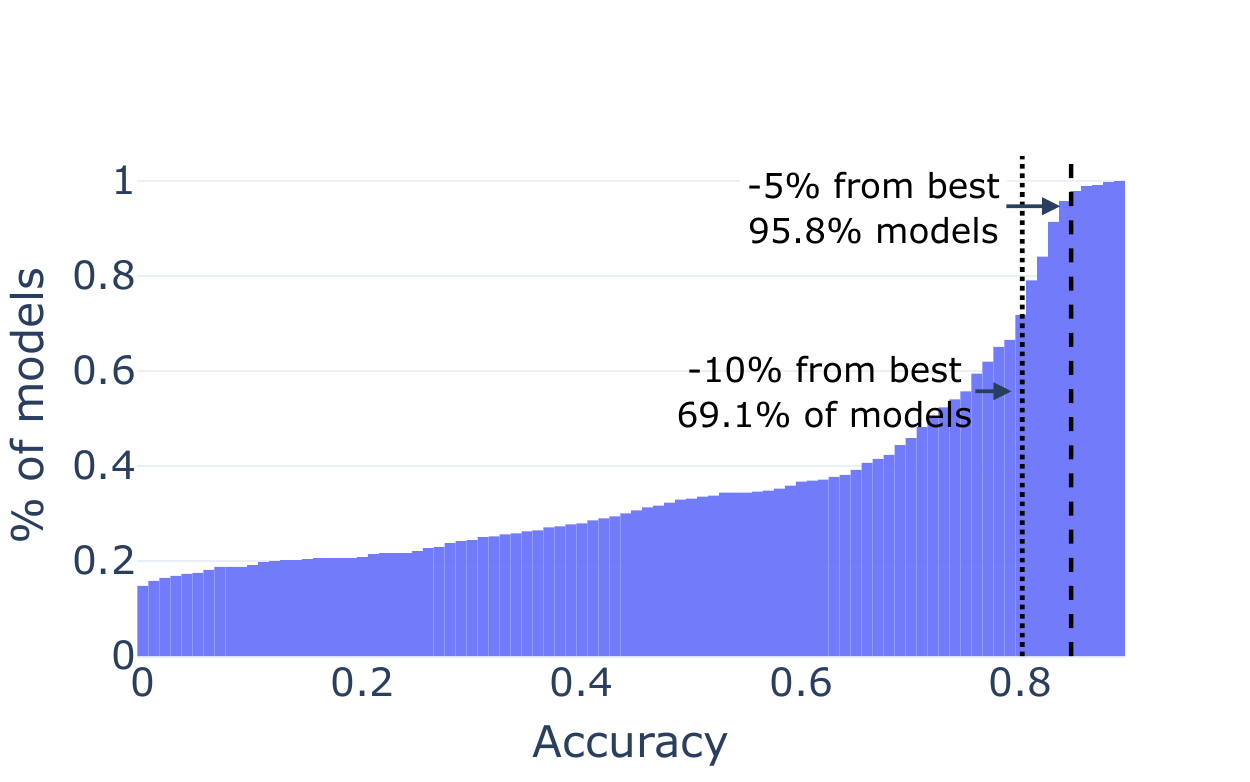} & 
        \includegraphics[width=0.31\linewidth]{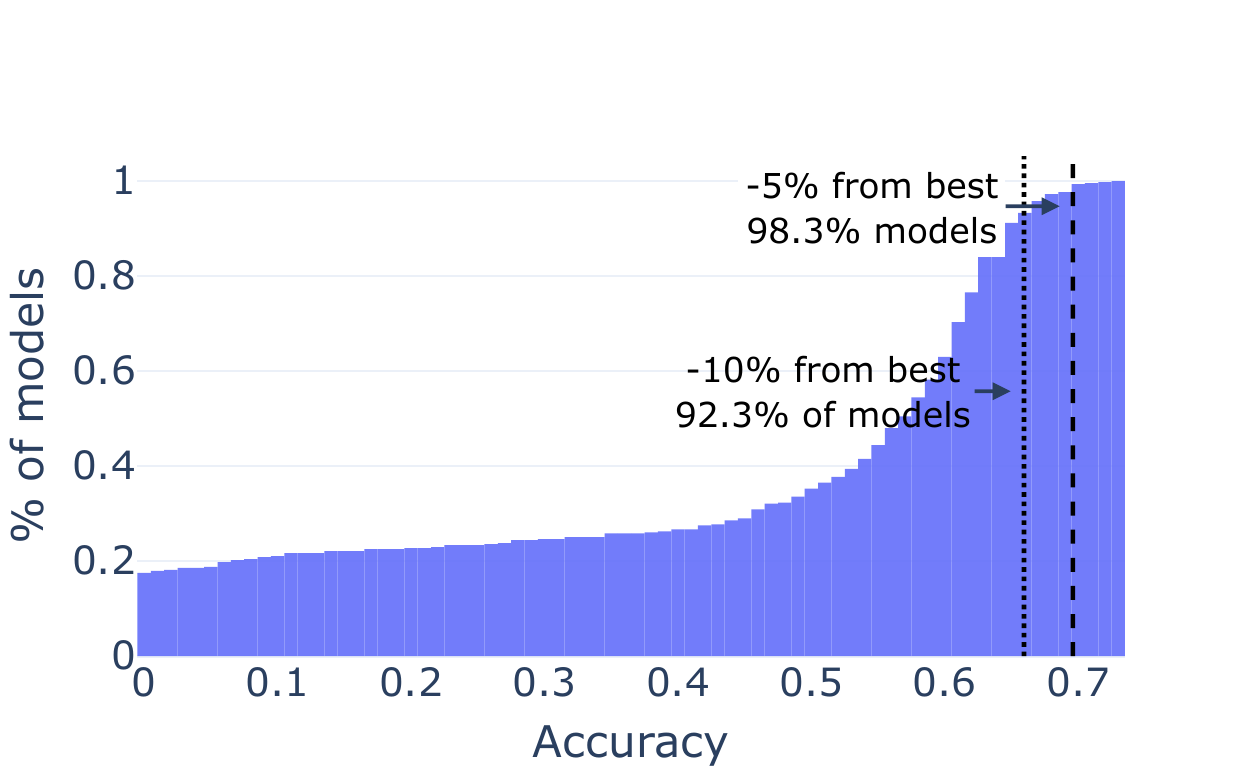} & 
        \includegraphics[width=0.31\linewidth]{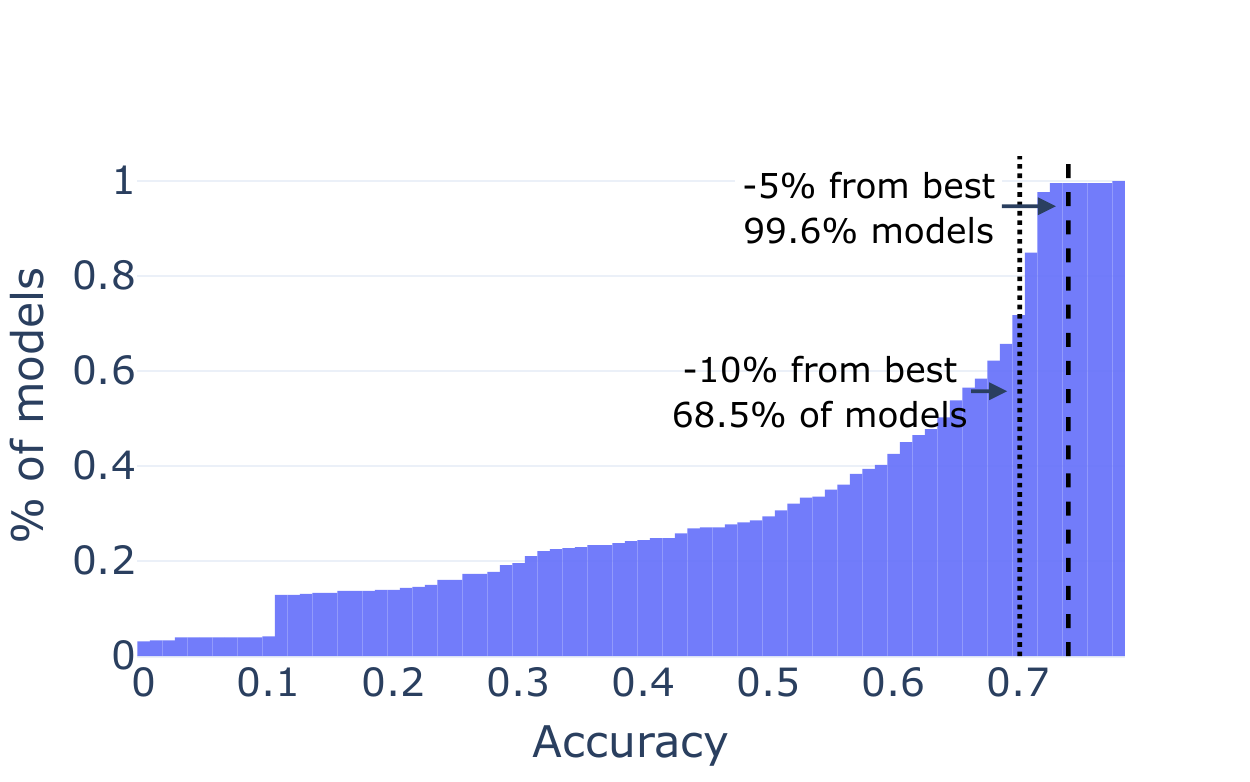} \\
        \shortstack{\small Qwen-3B \\ \textbf{GSM8K$_s$}} & 
        \shortstack{\small Qwen-3B \\ \textbf{MBPP$_s$}} & 
        \shortstack{\small Qwen-3B \\ \textbf{RouterBench$_s$}} \\[12pt]

        \includegraphics[width=0.31\linewidth]{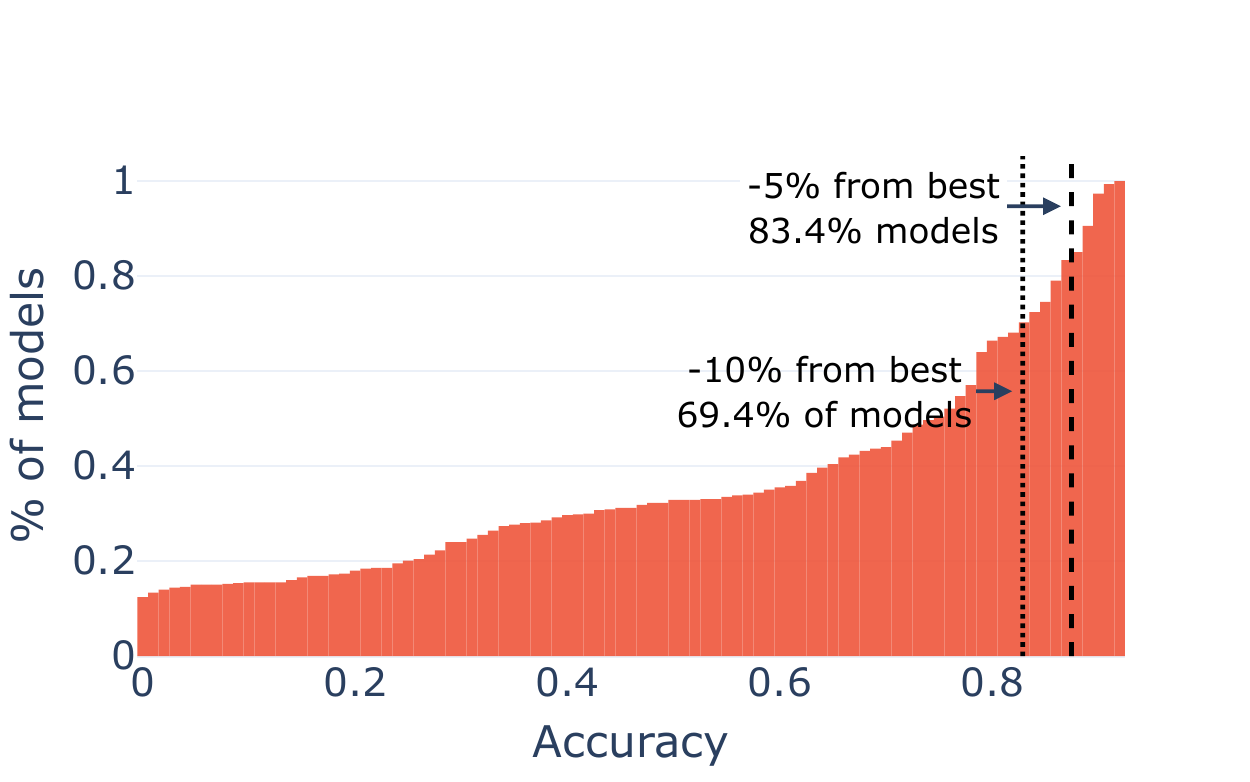} & 
        \includegraphics[width=0.31\linewidth]{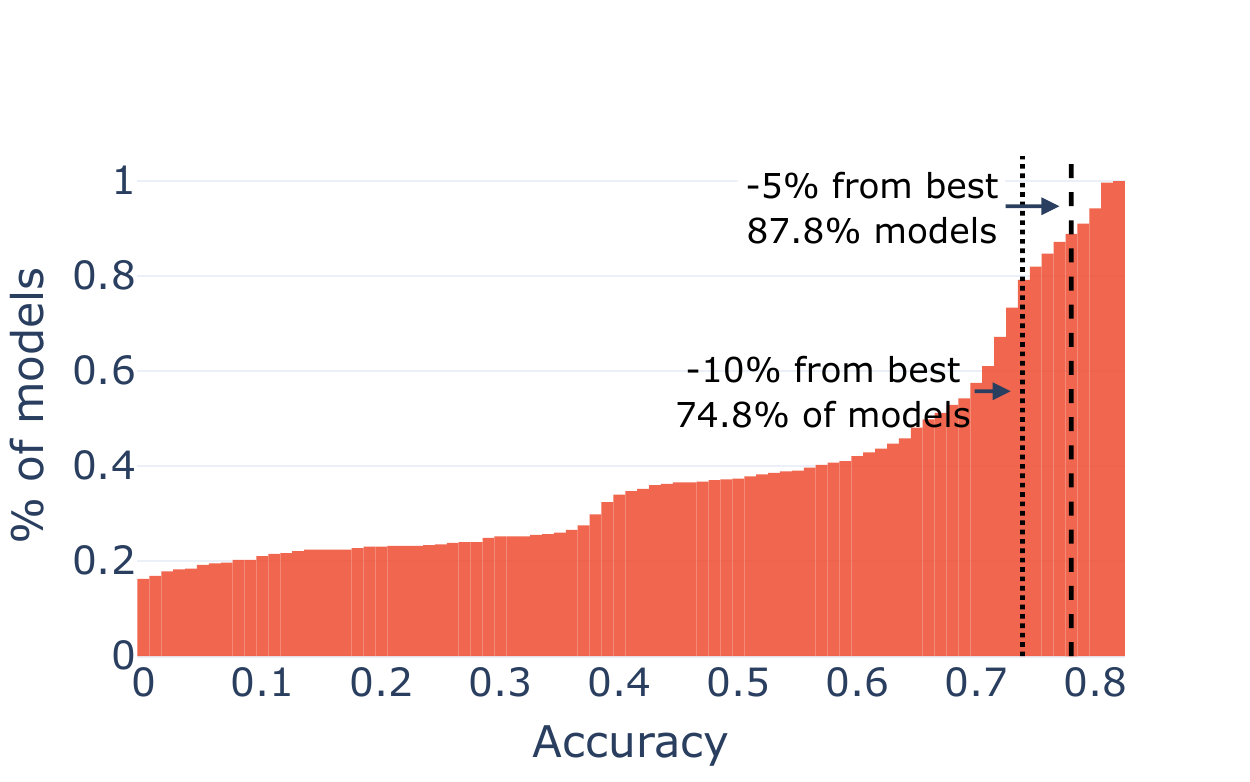} & 
        \includegraphics[width=0.31\linewidth]{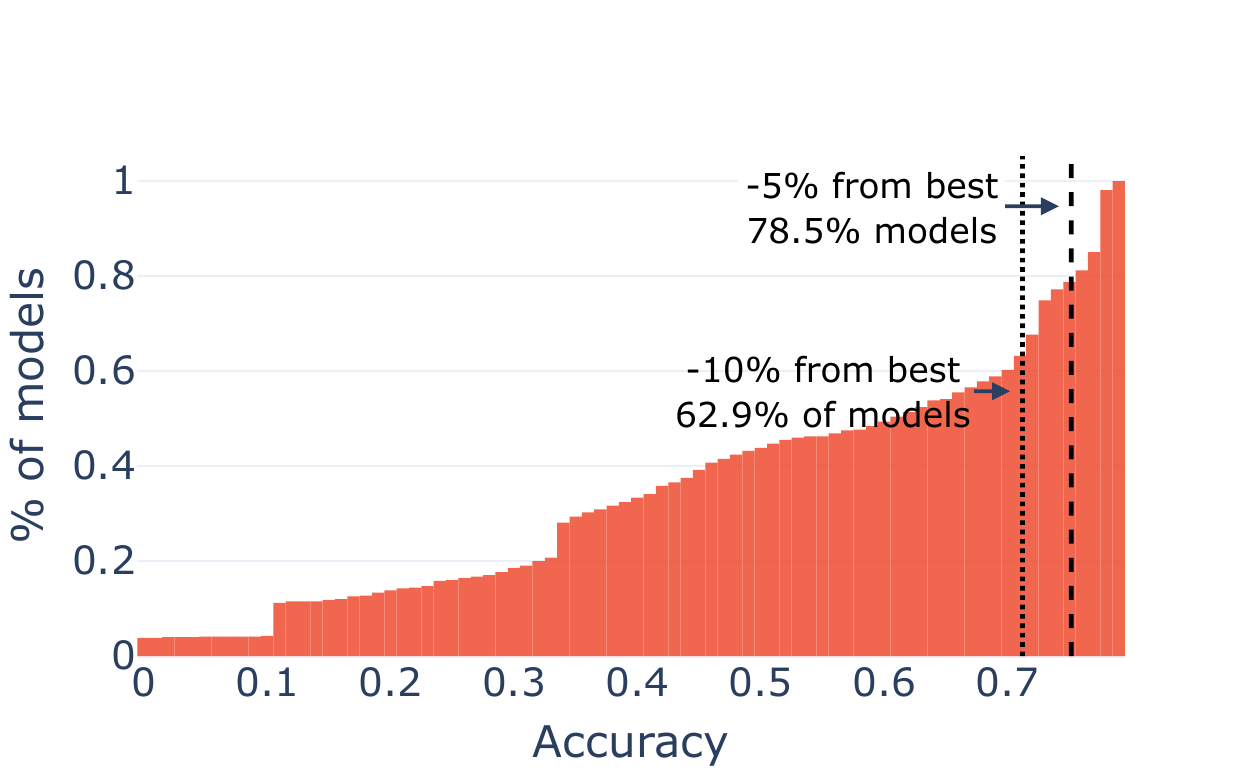} \\
        \shortstack{\small Qwen-7B \\ \textbf{GSM8K$_s$}} & 
        \shortstack{\small Qwen-7B \\ \textbf{MBPP$_s$}} & 
        \shortstack{\small Qwen-7B \\ \textbf{RouterBench$_s$}} \\[12pt]

        \includegraphics[width=0.31\linewidth]{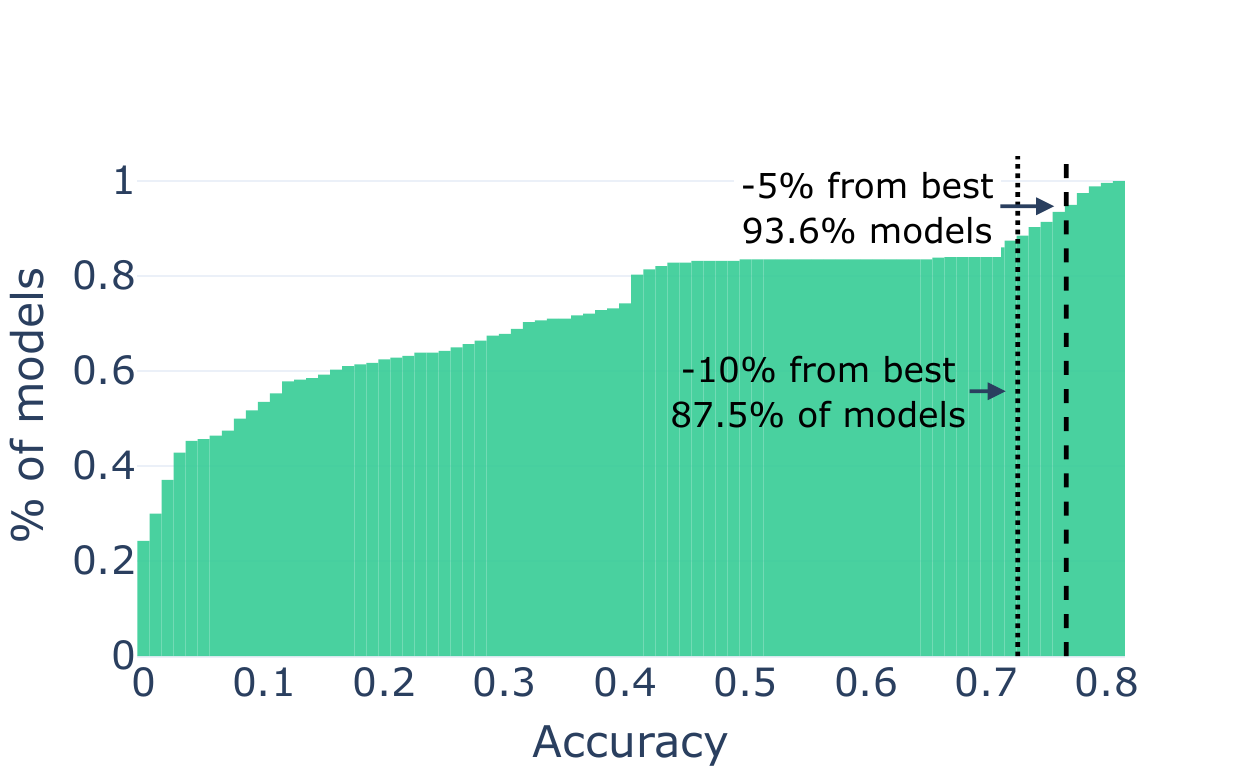} & 
        \includegraphics[width=0.31\linewidth]{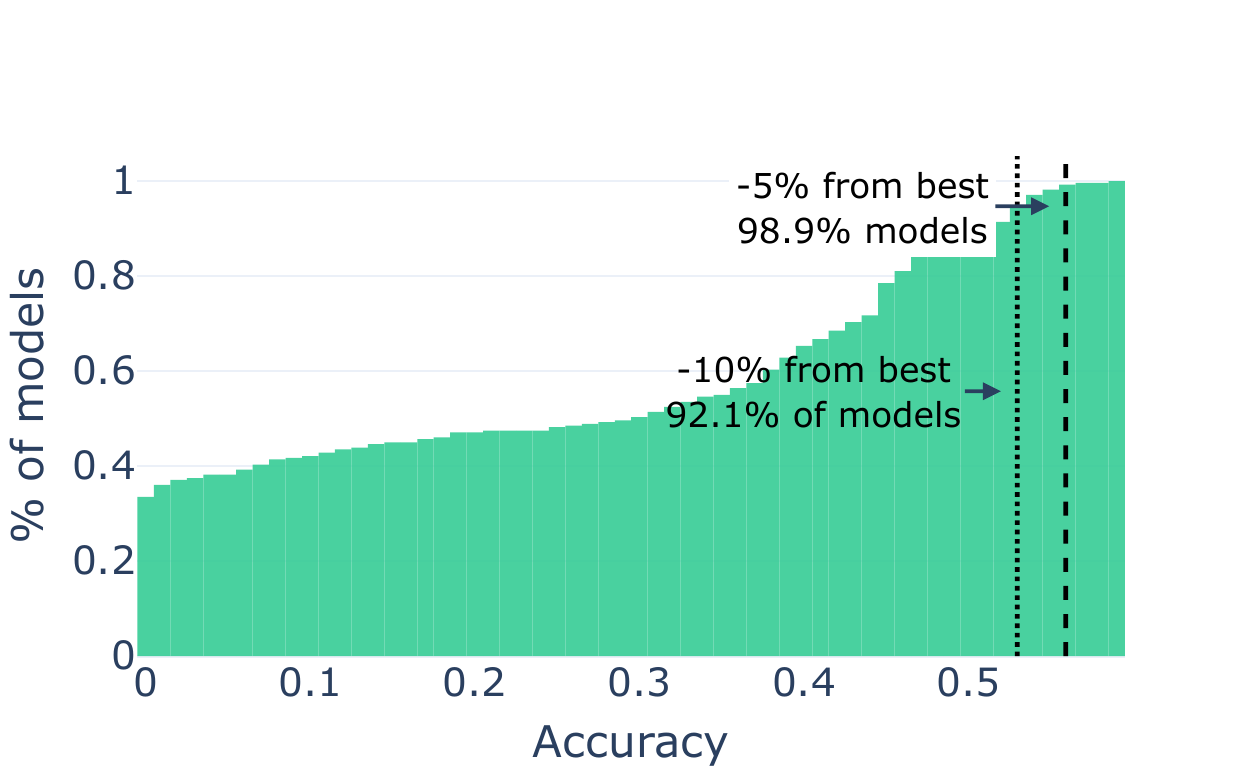} & 
        \includegraphics[width=0.31\linewidth]{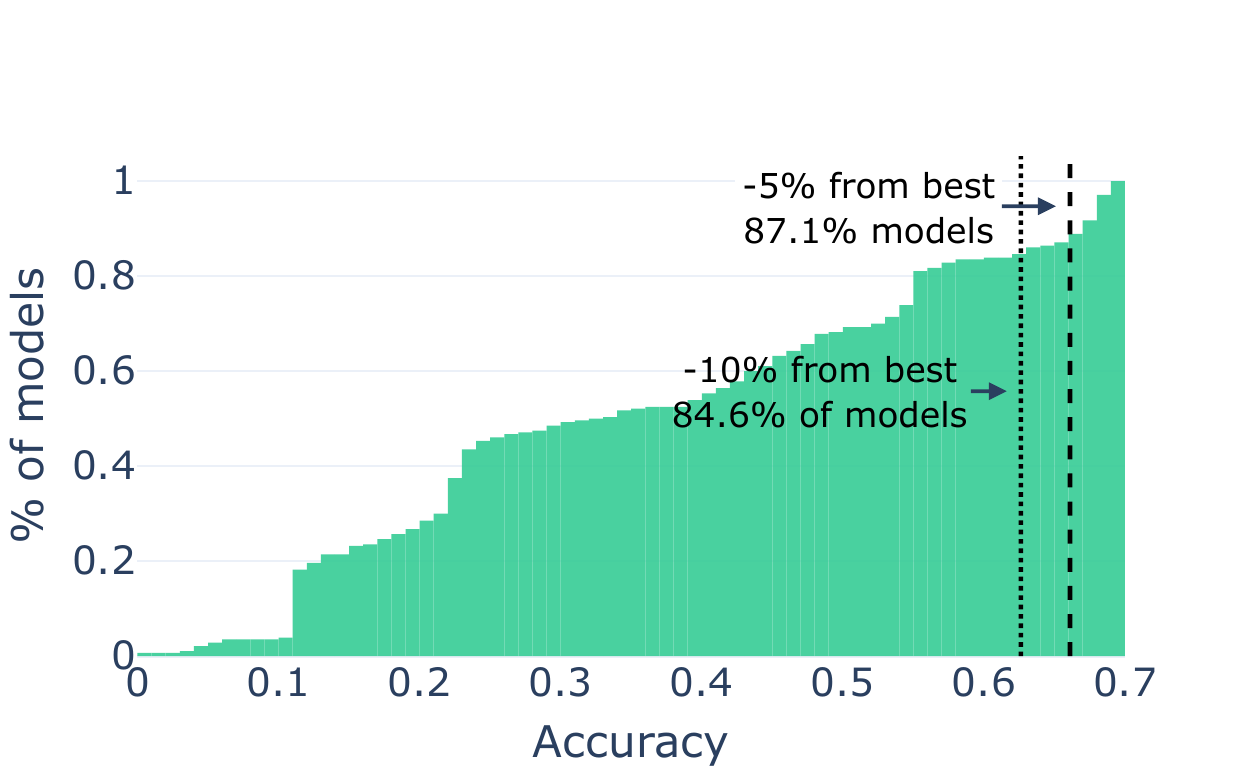} \\
        \shortstack{\small Mistral-7B \\ \textbf{GSM8K$_s$}} & 
        \shortstack{\small Mistral-7B \\ \textbf{MBPP$_s$}} & 
        \shortstack{\small Mistral-7B \\ \textbf{RouterBench$_s$}} \\[12pt]

        \includegraphics[width=0.31\linewidth]{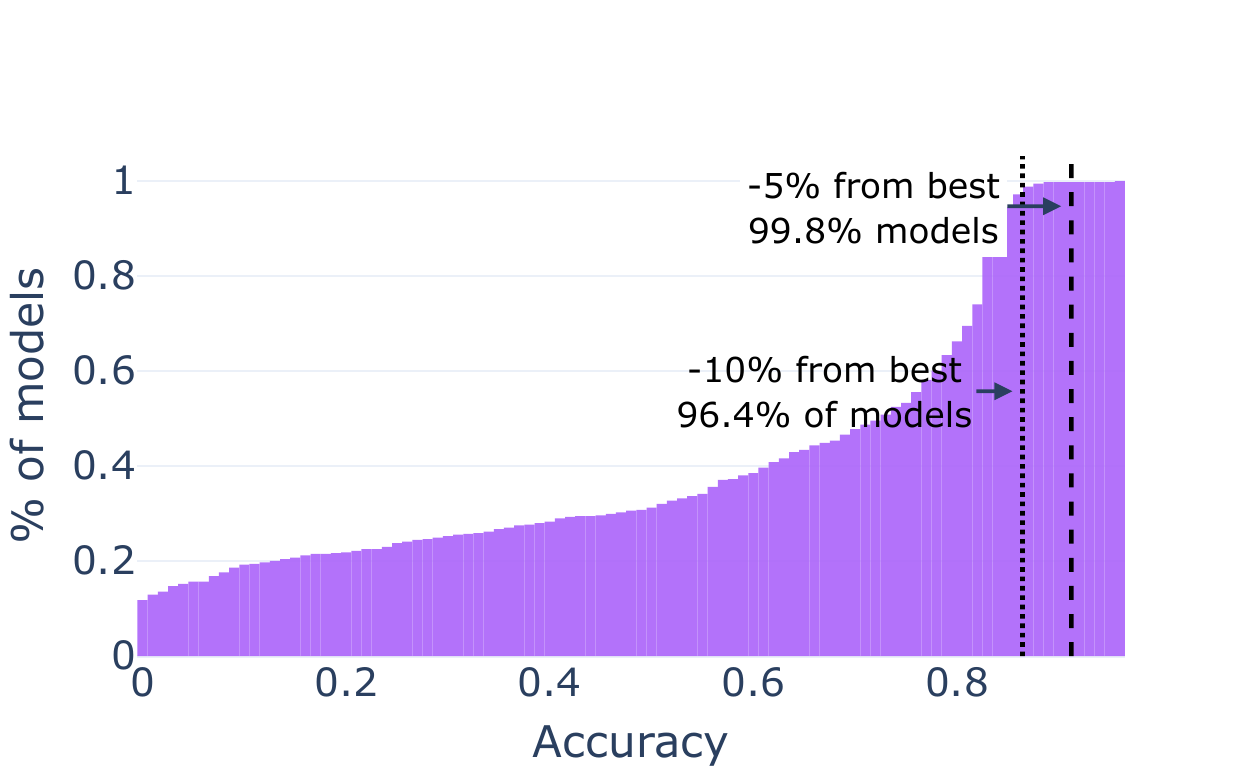} & 
        \includegraphics[width=0.31\linewidth]{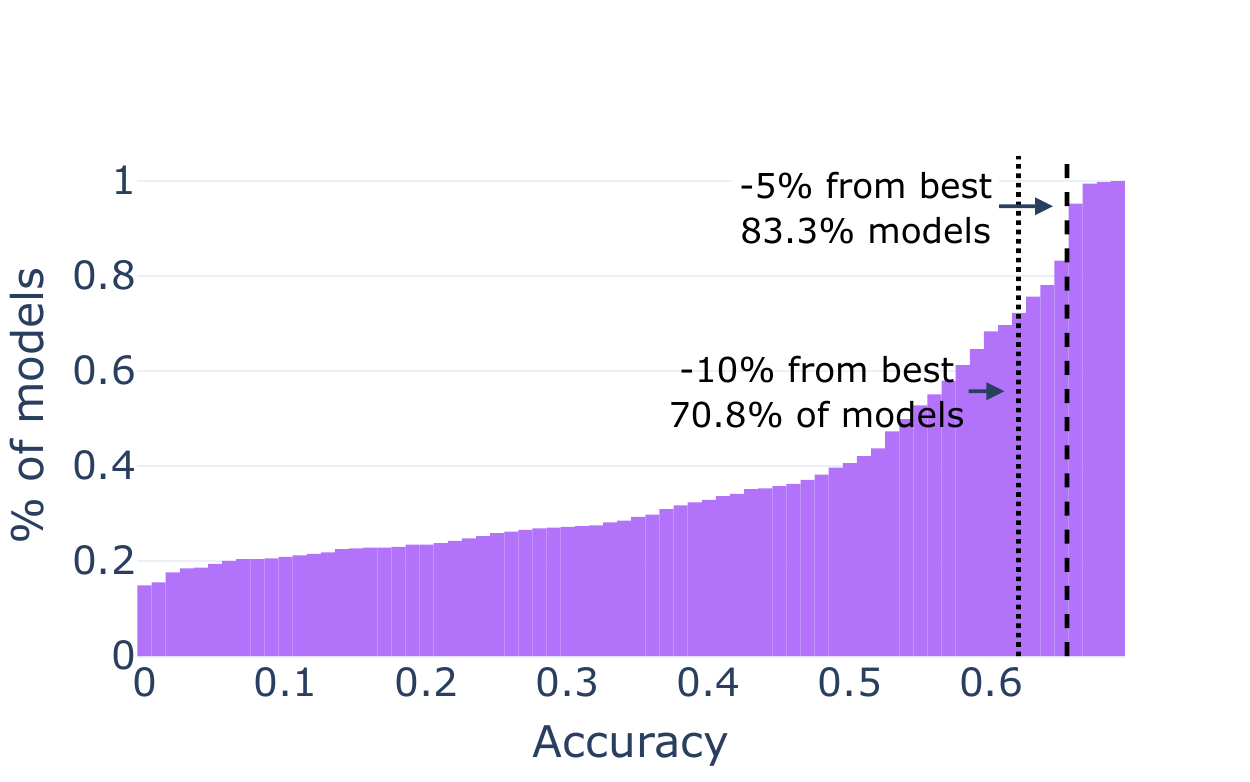} & 
        \includegraphics[width=0.31\linewidth]{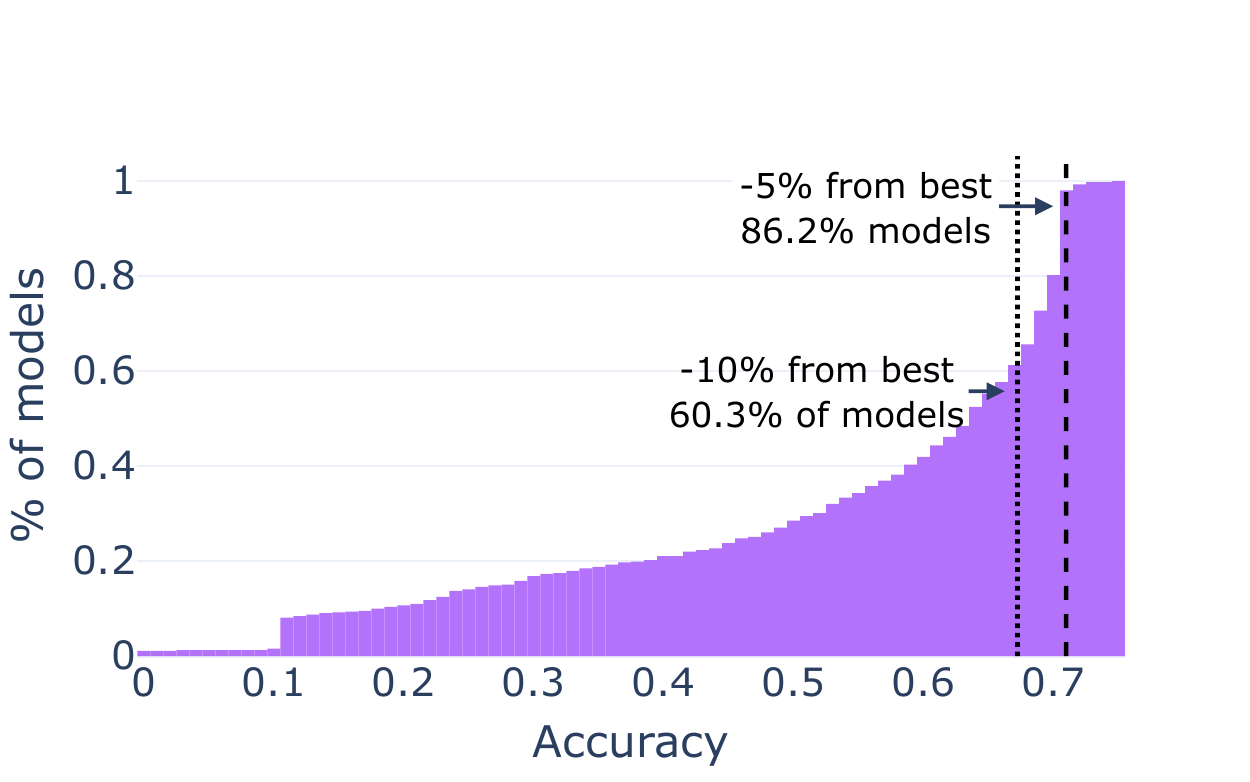} \\
        \shortstack{\small Llama3-8B \\ \textbf{GSM8K$_s$}} & 
        \shortstack{\small Llama3-8B \\ \textbf{MBPP$_s$}} & 
        \shortstack{\small Llama3-8B \\ \textbf{RouterBench$_s$}}
    \end{tabular}
    \caption{Cumulative Accuracy Distributions across different model architectures and specialized tasks.}
    \label{fig:distributions}
\end{figure*}

\section{Gem Documentation}
\label{app:documentation}
We analyze whether identified hidden gems include performance documentation. Results are show in Tab.~\ref{tab:hidden_gems_doc}.
We find that $19$ out of the $24$ gems has no performance documentation at all making them completely undetectable for text-based search. Moreover, out of the remaining $5$, $3$ of them are documentations targeted for South-Eastern Asian languages evaluation which are irrelevant for their gem tasks. This means only $2$ out of the $24$ gems could have been found by text search. Specifically, the first is Nvidia's official fine-tuning of Llama for math evaluated on the same evaluation dataset and the other model is a Qwen-3B fine-tuning which reports Fortran code evaluation, where we looked for Python coding skills. As many different coding models and evaluations exist, it is likely that text-based search engines could not have known this model is also the best available for Python tasks, and would not return it as their prime candidate.

\section{Additional Atlas Visualizations} 

\begin{figure*}[b]
    \centering
    \begin{tabular}{cc}
        \includegraphics[width=0.45\linewidth]{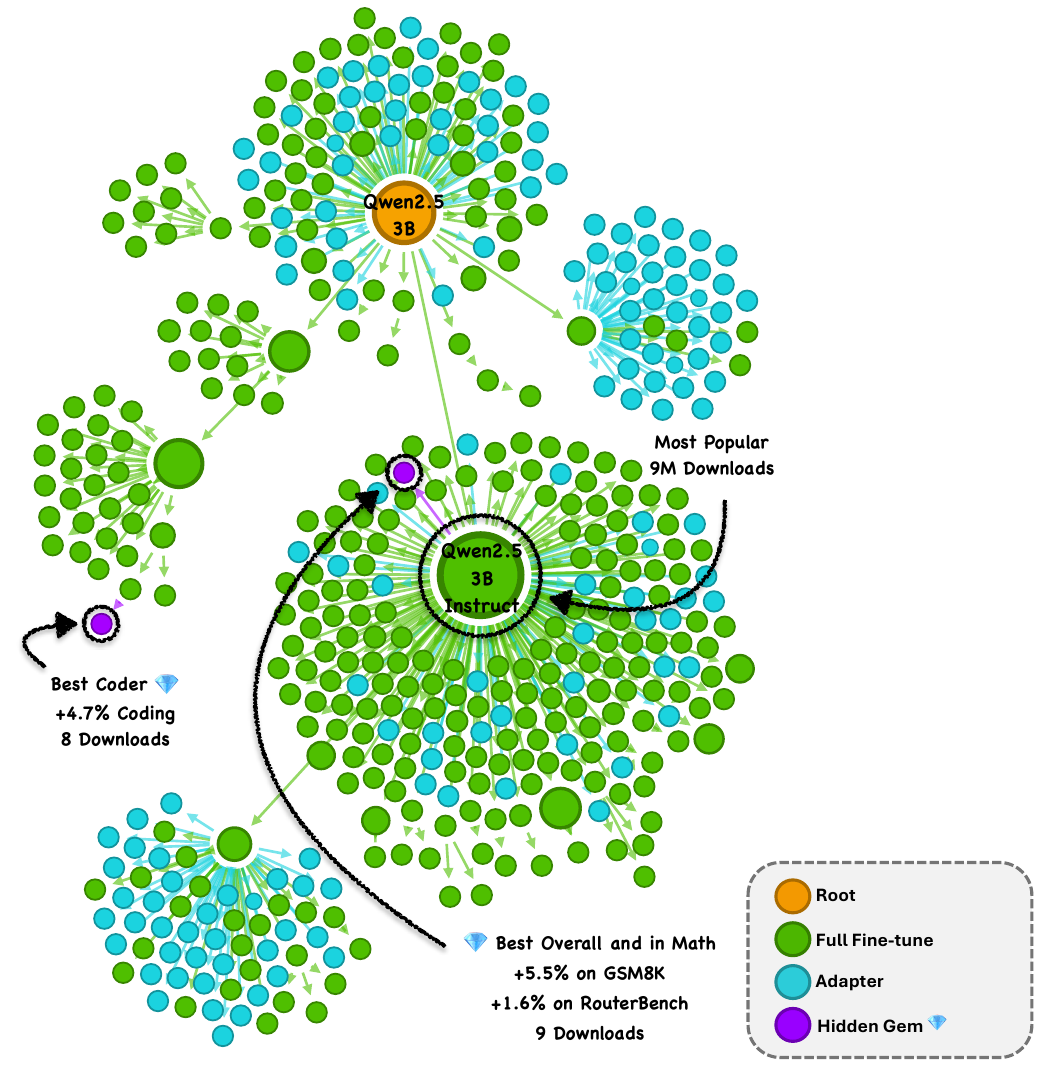} & 
        \includegraphics[width=0.45\linewidth]{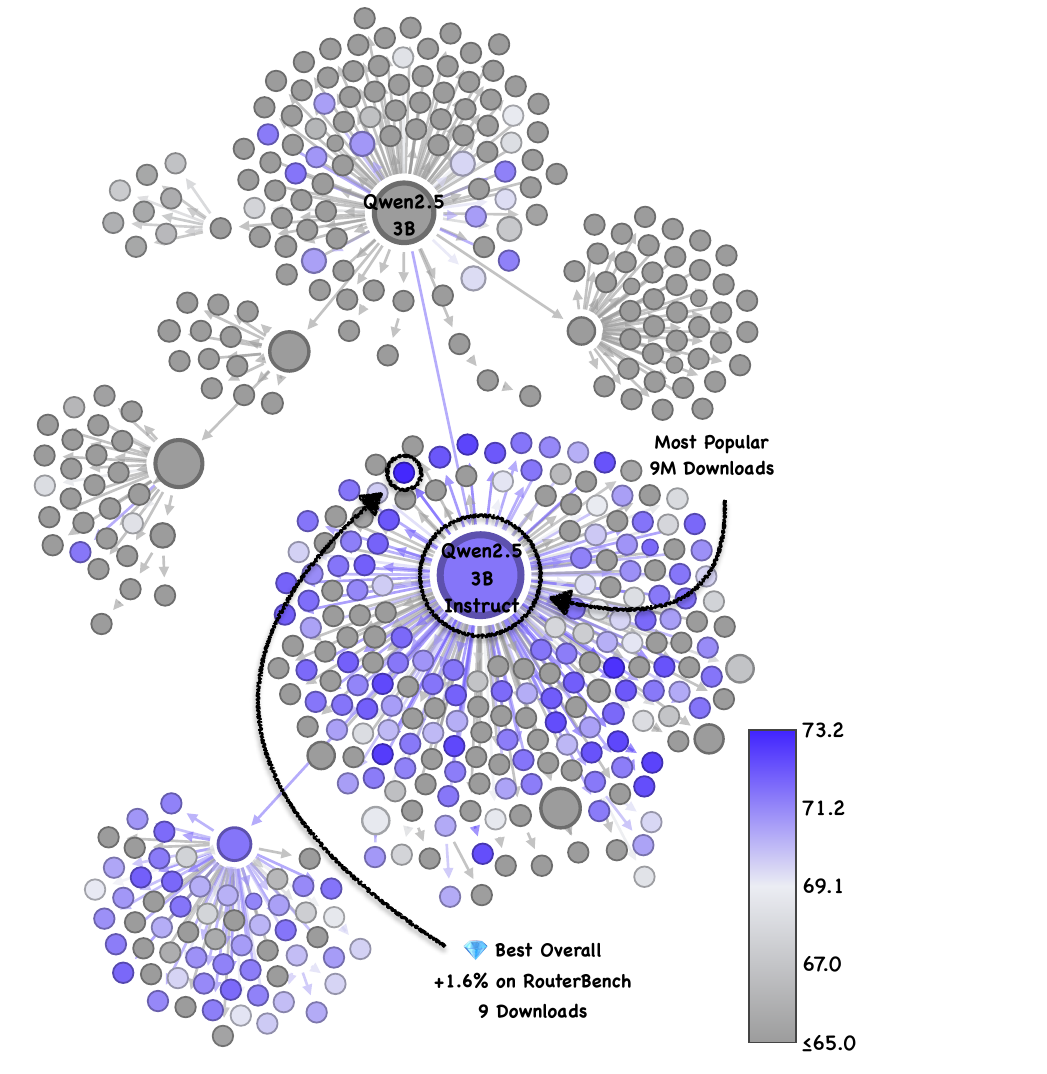} \\
        \textbf{\textit{Qwen-3B Model Tree}} & \textbf{\textit{RouterB.$_s$ Performance}} \\[12pt]
        
        \includegraphics[width=0.45\linewidth]{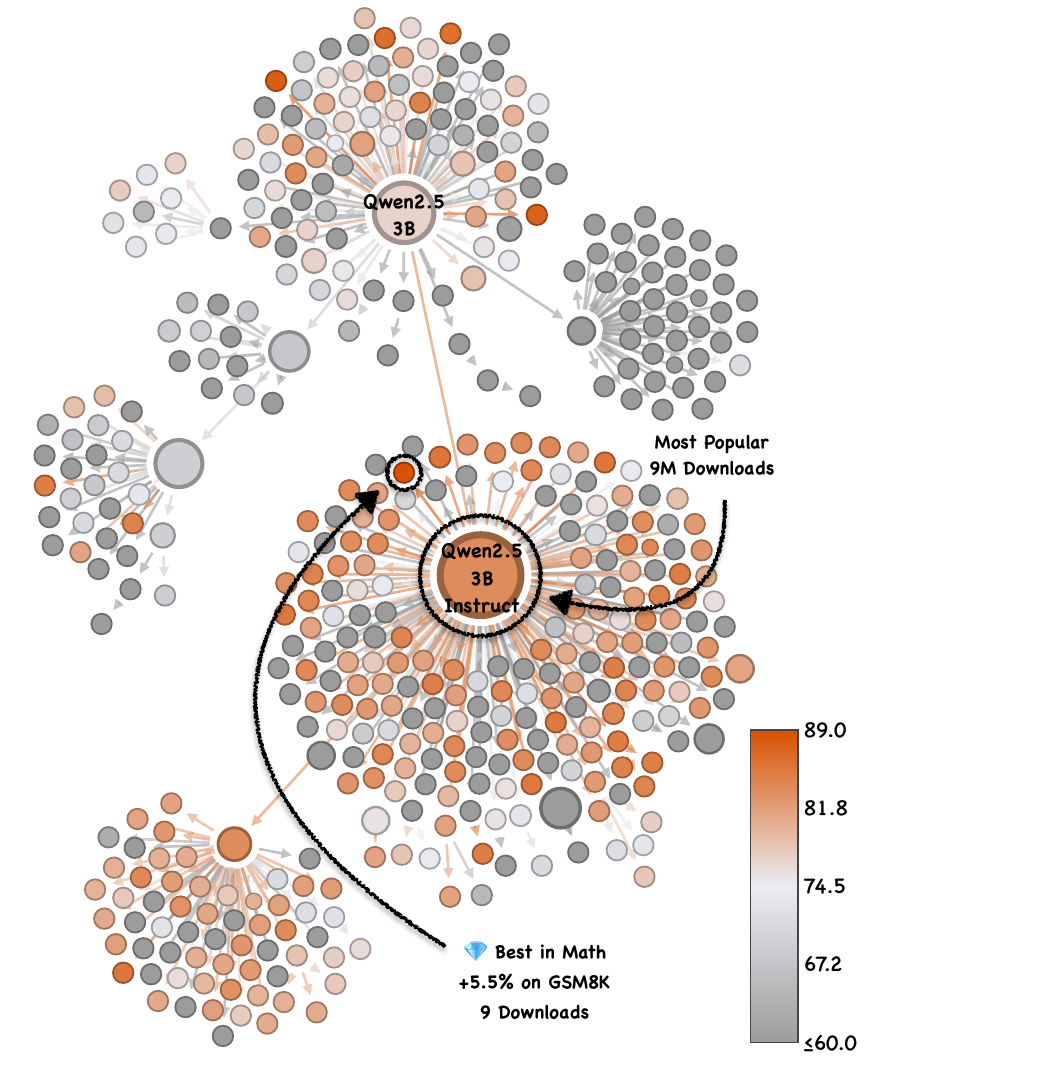} & 
        \includegraphics[width=0.45\linewidth]{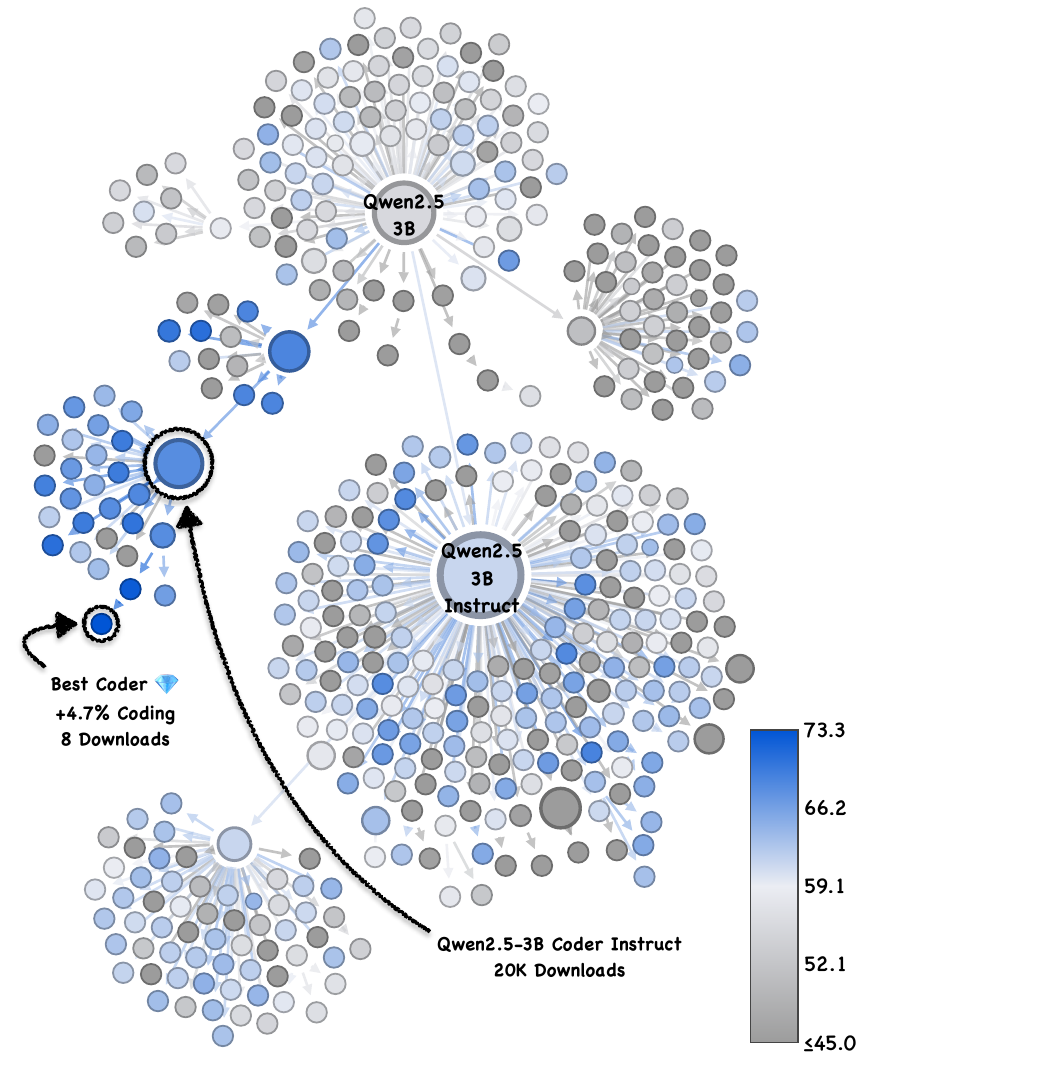} \\
        \textbf{\textit{GSM8K$_s$ Performance}} & \textbf{\textit{MBPP$_s$ Performance}}
    \end{tabular}
    \caption{\textbf{\textit{Different performance views for the Qwen-3B model tree.}}}
    \label{fig:qwen3b_views}
\end{figure*}

\begin{figure*}[b]
    \centering
    \begin{tabular}{cc}
        \includegraphics[width=0.45\linewidth]{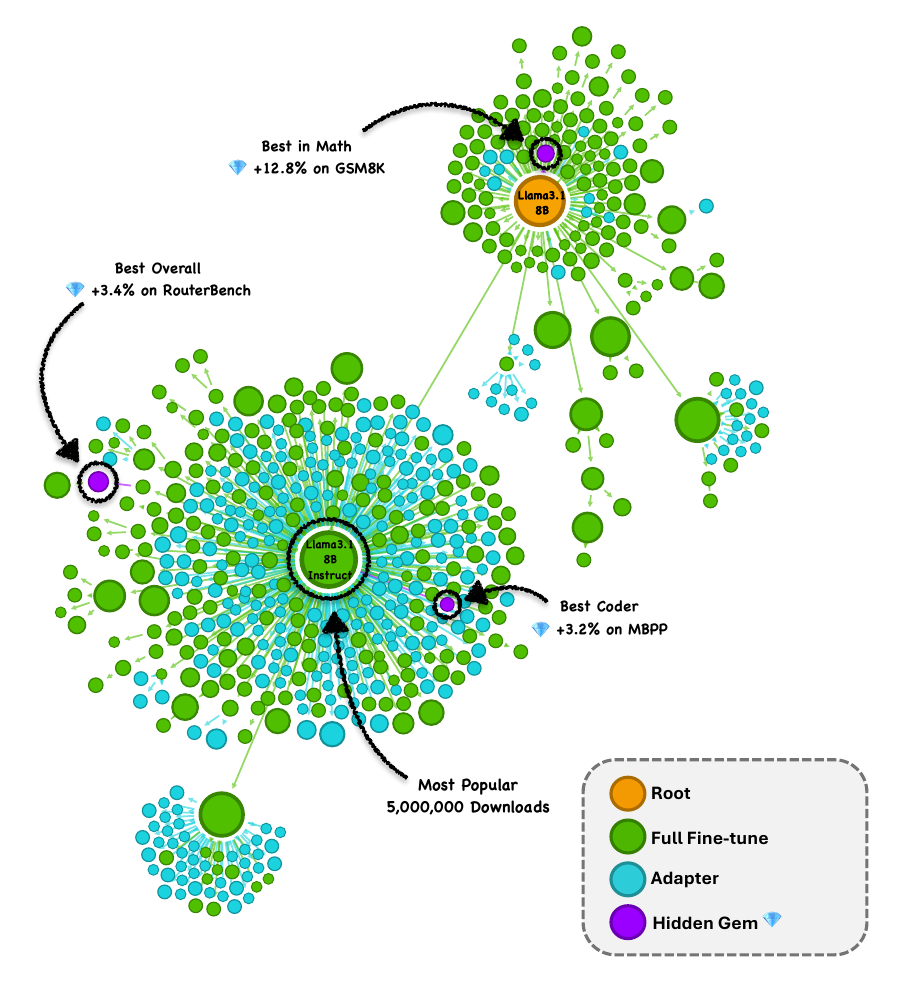} & 
        \includegraphics[width=0.45\linewidth]{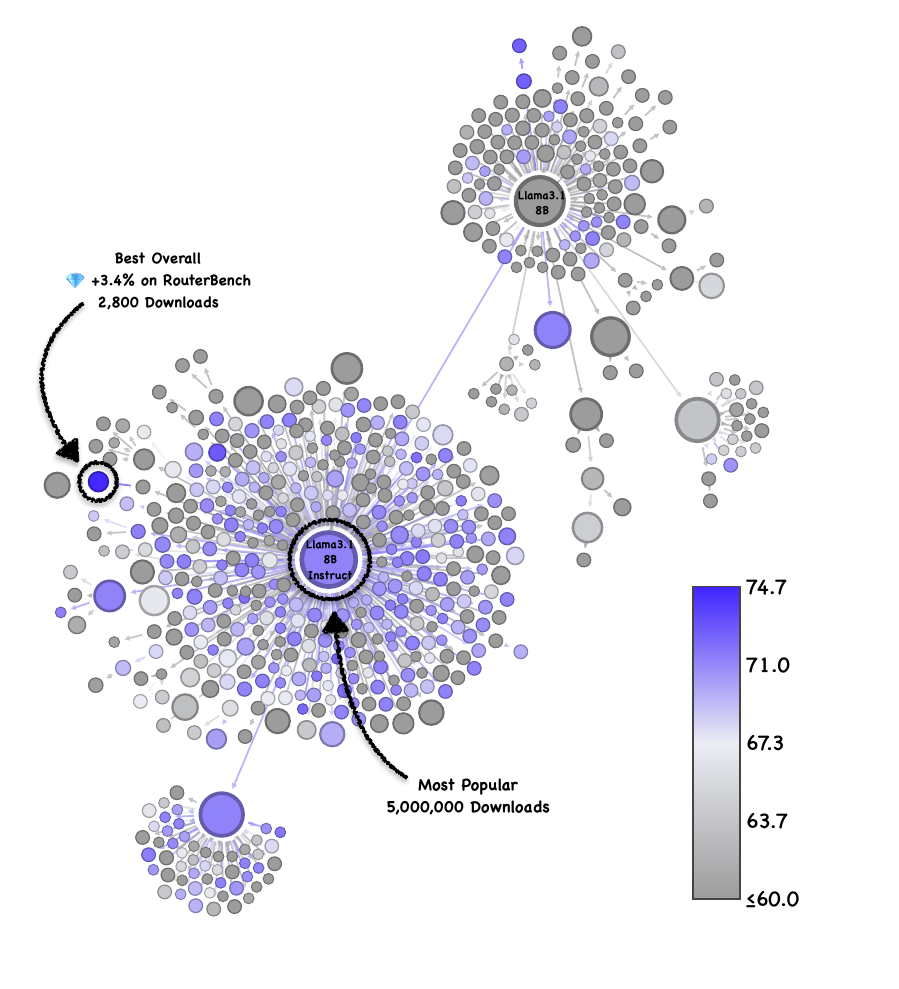} \\
        \textbf{\textit{Llama3-8B Model Tree}} & \textbf{\textit{RouterB.$_s$ Performance}} \\[12pt]
        
        \includegraphics[width=0.45\linewidth]{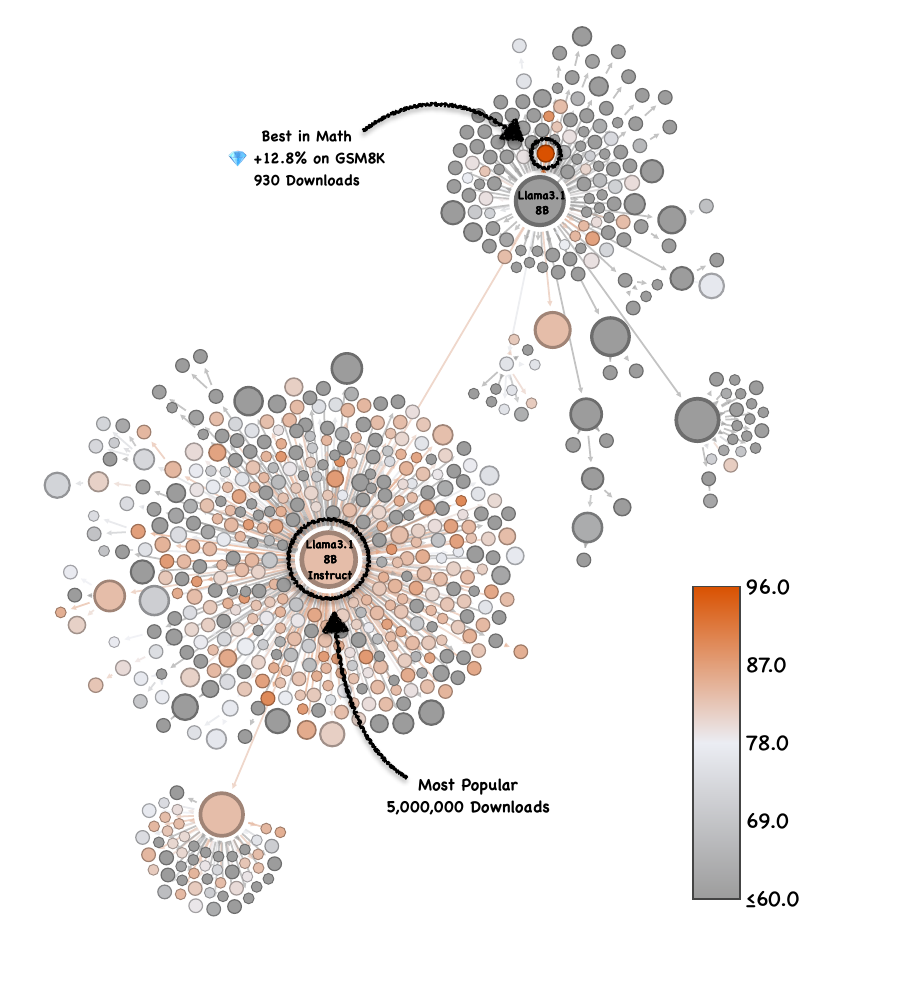} & 
        \includegraphics[width=0.45\linewidth]{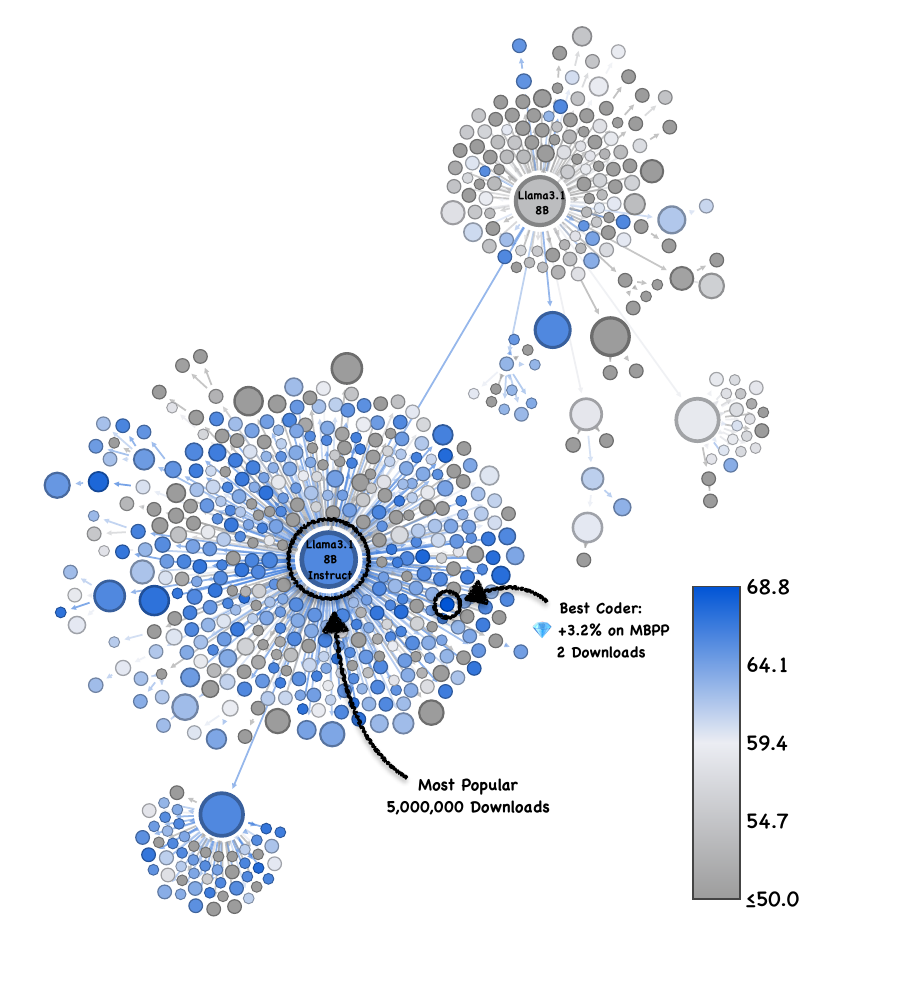} \\
        \textbf{\textit{GSM8K$_s$ Performance}} & \textbf{\textit{MBPP$_s$ Performance}}
    \end{tabular}
    \caption{\textbf{\textit{Different performance views for the Llama3-8B model tree.}}}
    \label{fig:llama3_views}
\end{figure*}

\begin{figure*}[b]
    \centering
    \begin{tabular}{cc}
        \includegraphics[width=0.45\linewidth]{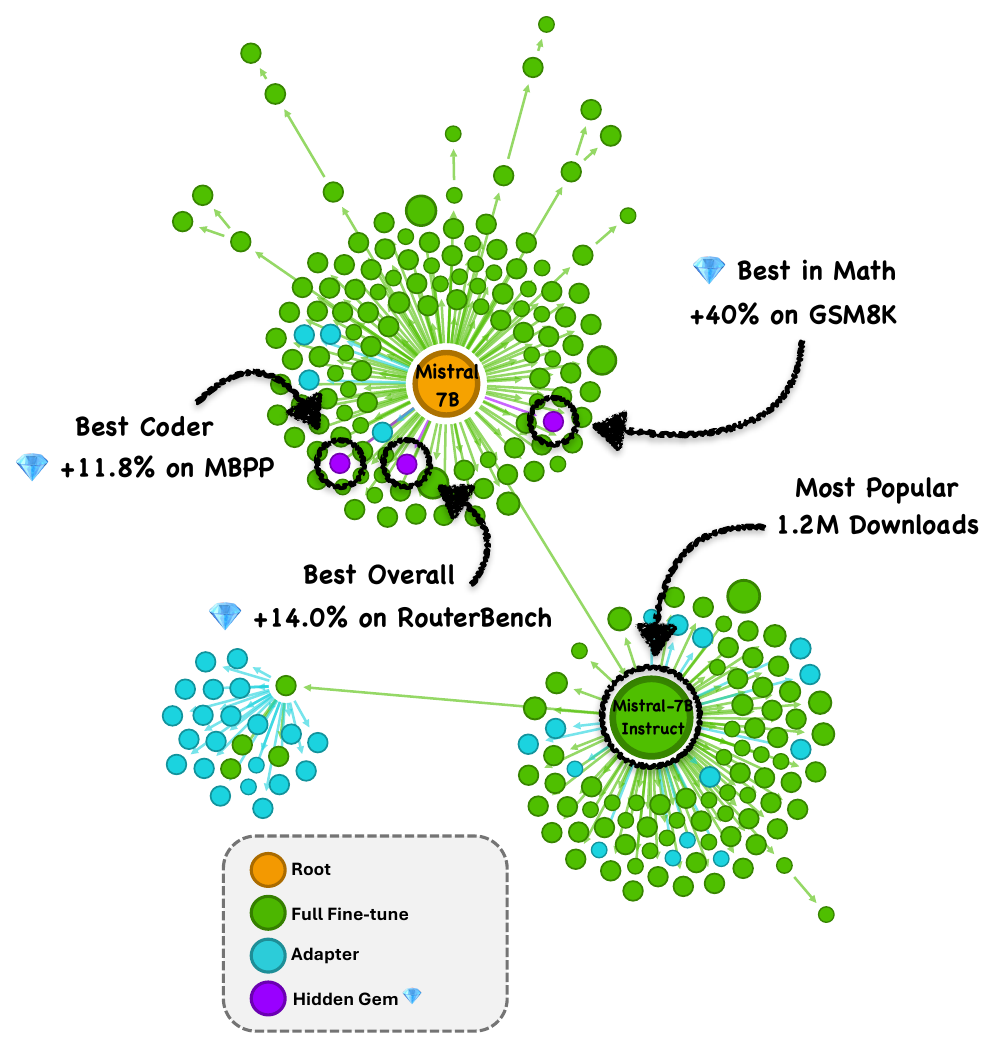} & 
        \includegraphics[width=0.45\linewidth]{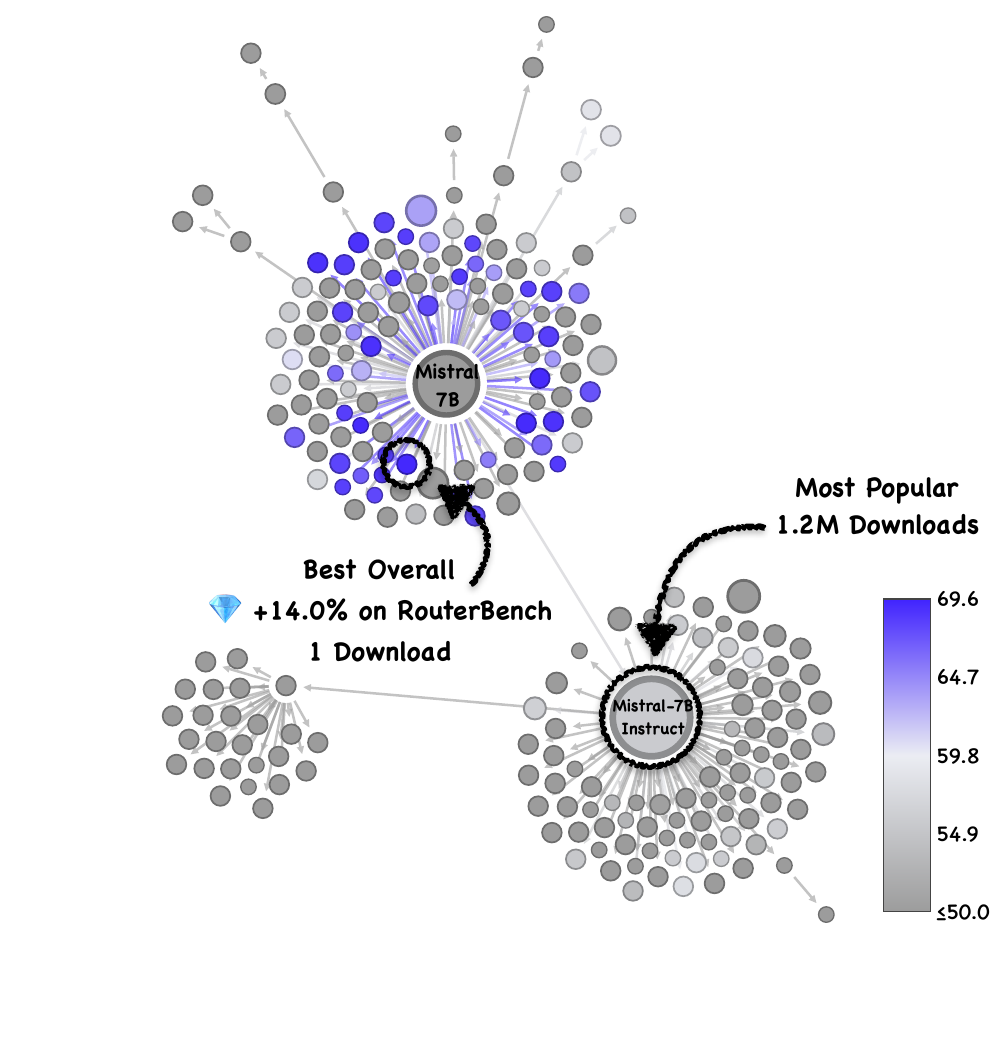} \\
        \textbf{\textit{Mistral-7B Model Tree}} & \textbf{\textit{RouterB.$_s$ Performance}} \\[12pt]
        
        \includegraphics[width=0.45\linewidth]{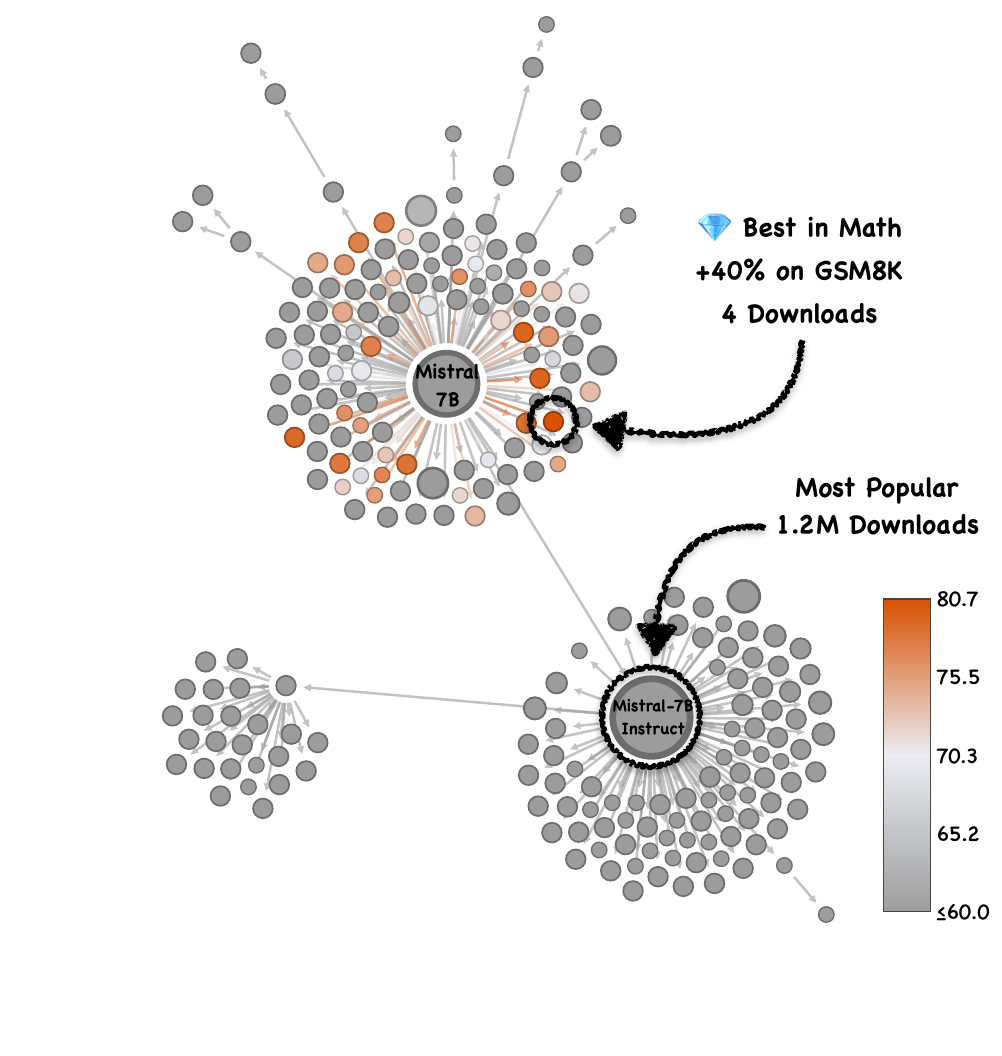} & 
        \includegraphics[width=0.45\linewidth]{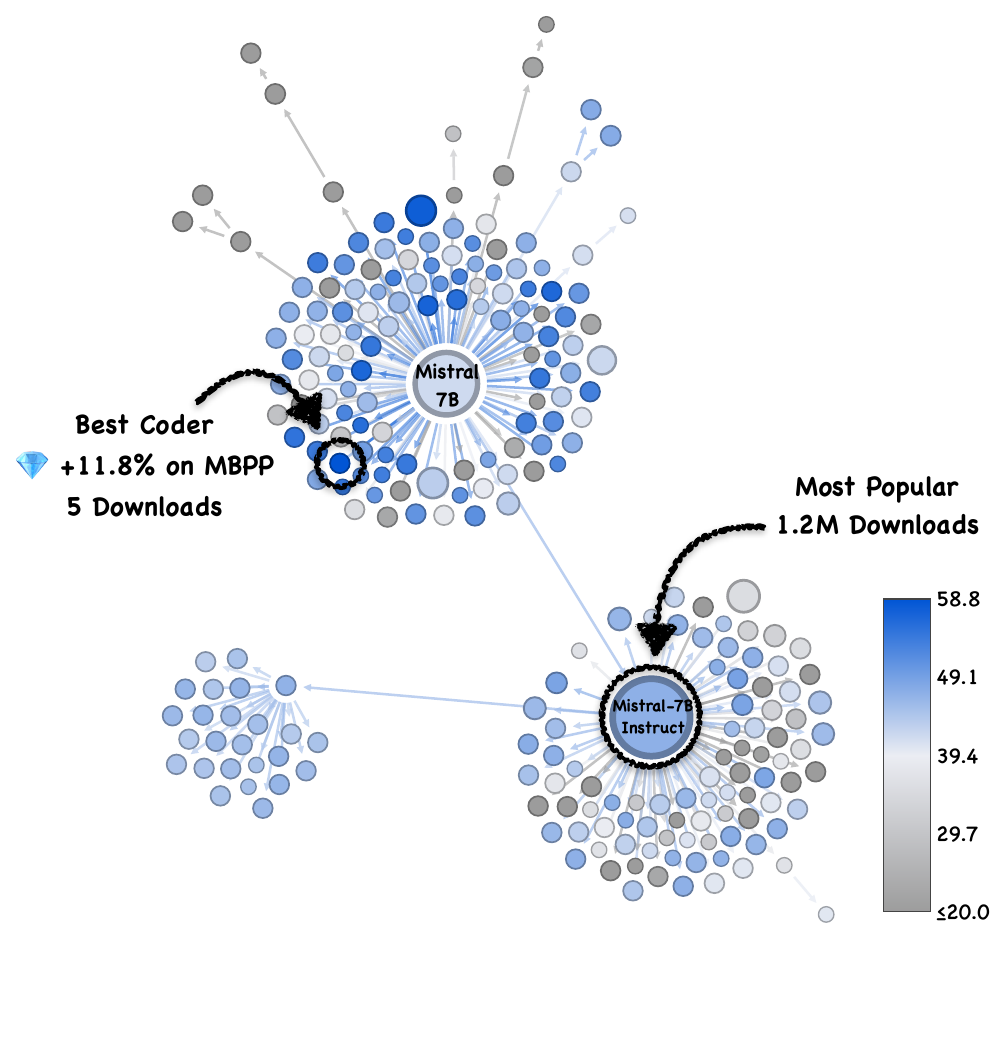} \\
        \textbf{\textit{GSM8K$_s$ Performance}} & \textbf{\textit{MBPP$_s$ Performance}}
    \end{tabular}
    \caption{\textbf{\textit{Different performance views for the Mistral-7B model tree.}}}
    \label{fig:mistral7b_views}
\end{figure*}

\begin{figure*}[t]
    \centering
    \begin{tabular}{cc}
        \includegraphics[width=0.45\linewidth]{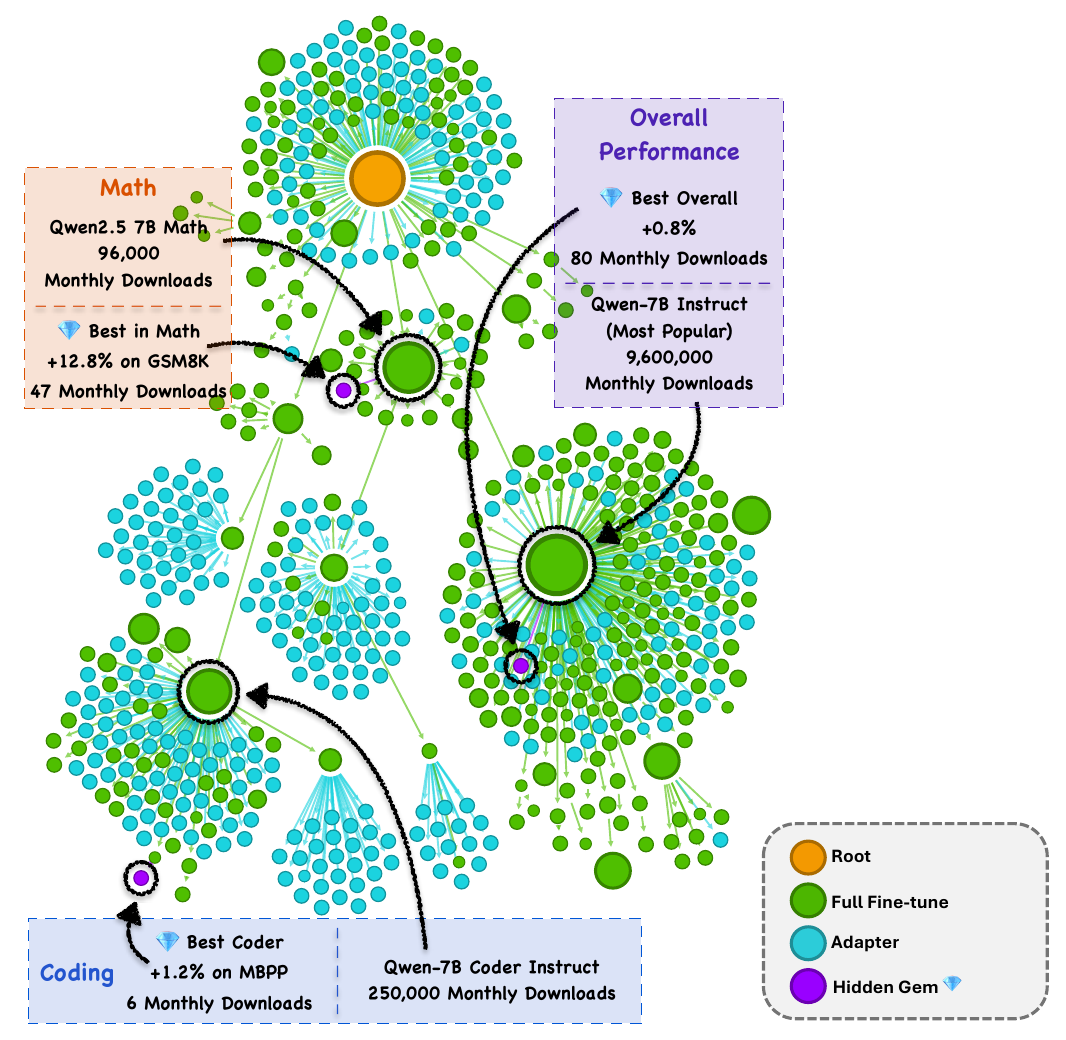} & 
        \includegraphics[width=0.45\linewidth]{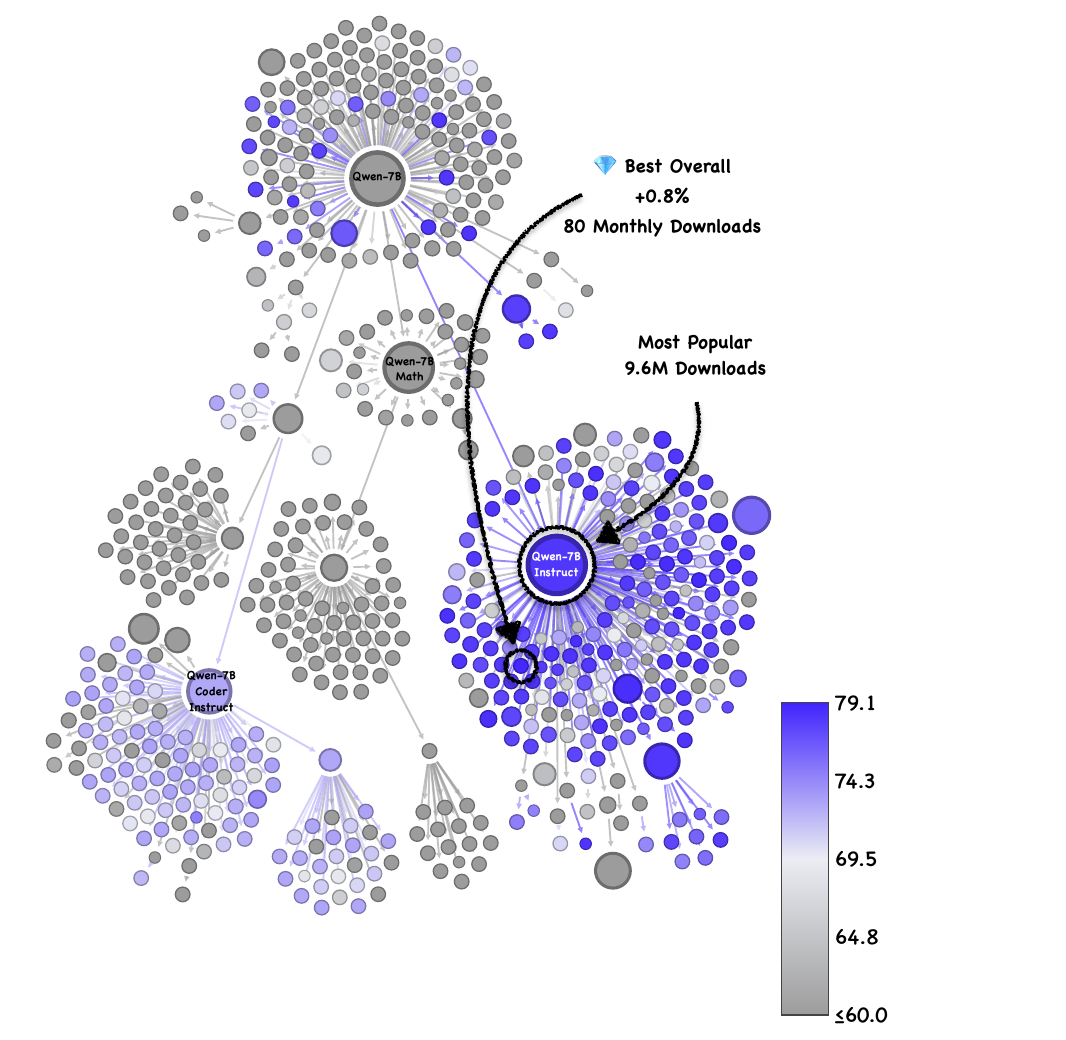} \\
        \textbf{\textit{Qwen-7B Model Tree}} & \textbf{\textit{RouterB.$_s$ Performance}} \\[12pt]
        
        \includegraphics[width=0.45\linewidth]{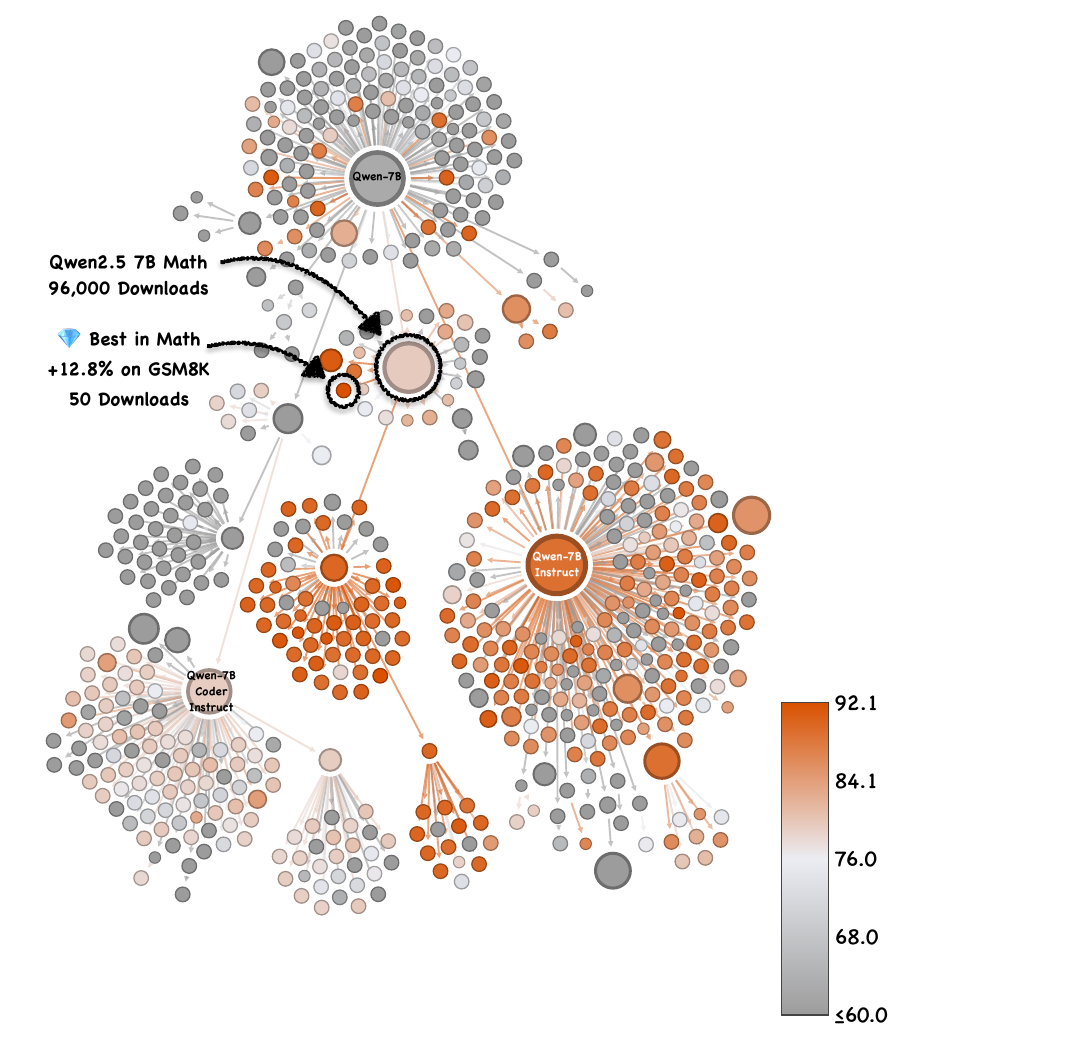} & 
        \includegraphics[width=0.45\linewidth]{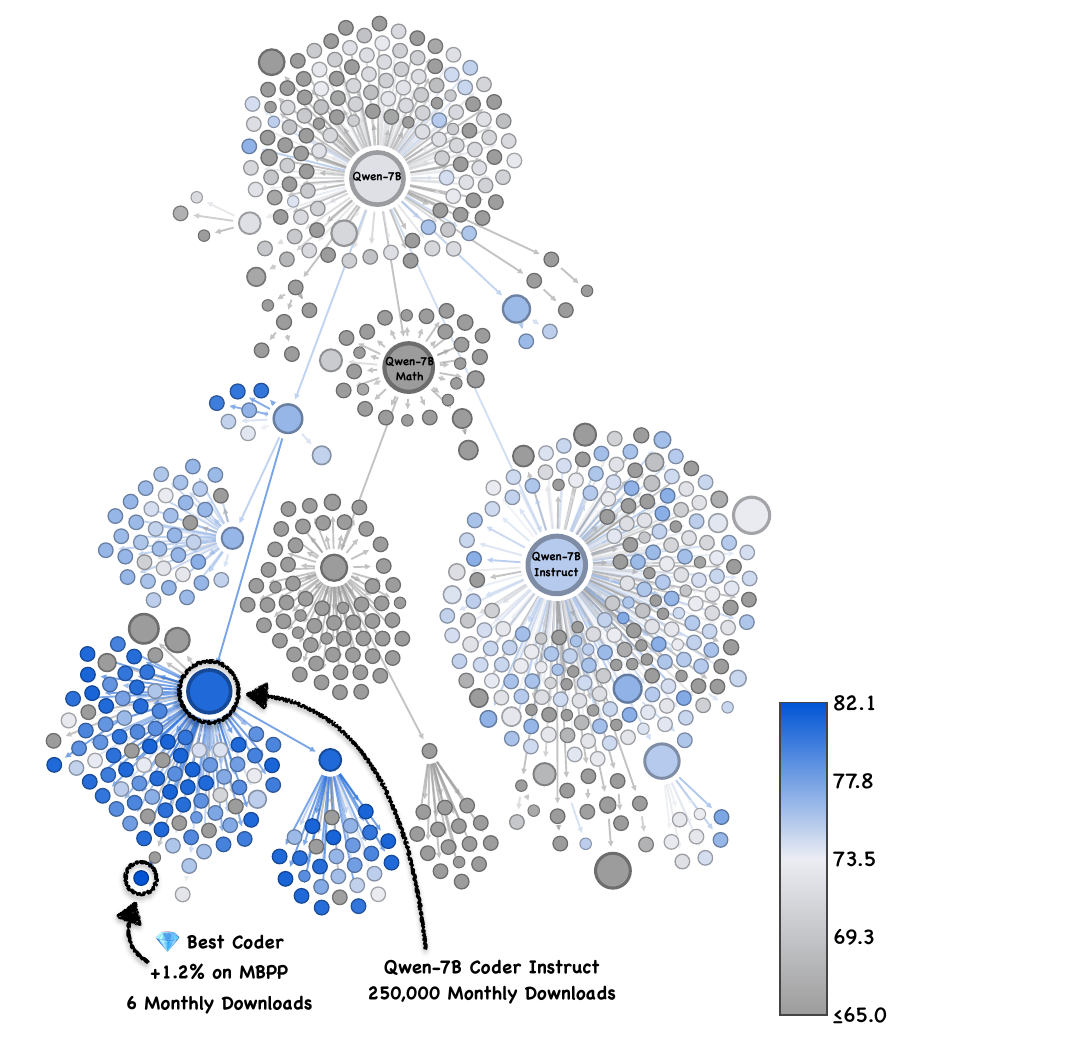} \\
        \textbf{\textit{GSM8K$_s$ Performance}} & \textbf{\textit{MBPP$_s$ Performance}}
    \end{tabular}
    \caption{\textbf{\textit{Different performance views for the Qwen-7B model tree.}}}
    \label{fig:qwen7b_views}
\end{figure*}

In Fig.~\ref{fig:qwen3b_views}--\ref{fig:qwen7b_views}, we provide extended atlas visualizations for the selected model trees, color-coded by performance on GSM8K$_s$ (math), MBPP$_s$ (coding), and RouterBench$_s$ (overall). These views again show that \q{hidden gems} often possess limited download counts compared to the popular base versions, making them nearly invisible when observing the tree's structure alone. Furthermore, gem models are often scattered across the tree rather than clustered in predictable branches. A notable exception is the Qwen-7B tree, where high-performers are nested within relevant official fine-tuning branches. For instance, the top coding models reside within the Qwen-7B Coder Instruct sub-tree. However, such curated official versions are not universally available and even when present, their sub-trees can be very large, making an exhaustive search computationally prohibitive.

\clearpage

\begin{table*}[t!]
    \caption{\textbf{\textit{Qwen-3B Hidden Gems.}}}
    \vspace{0.2cm}
    \centering

    {
    \small
    \setlength{\tabcolsep}{8pt}
    \renewcommand{\arraystretch}{1.05}
    
    \begin{tabular}{ccccc}
        \textbf{Tree} & \textbf{Task} & \textbf{Best Found Performer} & \textbf{Accuracy} \\
        \toprule
        \multirow{3}{*}{\textbf{Qwen-3B}} & Coding & \texttt{GiuLeo01/FortranCodeGen-3B-SynthData} & 73.3 \\
        & Math & \texttt{watermelonhjg/Qwen2.5-3B-Instruct-EN-Zero} & 89.0 \\
        & General & \texttt{watermelonhjg/Qwen2.5-3B-Instruct-EN-Zero} & 73.2 \\
        \bottomrule
    \end{tabular}
    }
    \label{tab:qwen3b_gems}
\end{table*}

\clearpage

\begin{table*}[t!]
    \caption{\textbf{\textit{Llama3.1-8B Hidden Gems.}}}
    \vspace{0.2cm}
    \centering

    {
    \small
    \setlength{\tabcolsep}{8pt}
    \renewcommand{\arraystretch}{1.05}
    
    \begin{tabular}{ccccc}
        \textbf{Tree} & \textbf{Task} & \textbf{Best Found Performer} & \textbf{Accuracy} \\
        \toprule
        \multirow{3}{*}{\textbf{Llama3.1-8B}} & Coding & \texttt{plandes/sdoh-llama-3-1-8b} & 68.8 \\
        & Math & \texttt{nvidia/OpenMath2-Llama3.1-8B} & 96.0 \\
        & General & \texttt{aisingapore/Llama-SEA-LION-v3-8B-IT} & 74.7 \\
        \bottomrule
    \end{tabular}
    }
    \label{tab:llama_gems}
\end{table*}

\clearpage

\begin{table*}[t!]
    \caption{\textbf{\textit{Mistral-7B Hidden Gems.}}}
    \vspace{0.2cm}
    \centering
    {
    \small
    \setlength{\tabcolsep}{8pt}
    \renewcommand{\arraystretch}{1.5} 
    
    \begin{tabular}{llp{7.5cm}c}
        \textbf{Tree} & \textbf{Task} & \textbf{Best Found Performer} & \textbf{Accuracy} \\
        \toprule
        & \multirow{2}{*}{Coding} & \texttt{mlfoundations-dev/mistral\_7b\_0-3\_oh-} \newline \texttt{dcft-v3.1-claude-3-5-sonnet-20241022} & 58.8 \\
        \multirow{2}{*}{\textbf{Mistral-7B}} & \multirow{2}{*}{Math} & \texttt{mlfoundations-dev/oh-mistral-bs512\_lr5\_00E-06\_} \newline \texttt{schedulercosine\_with\_min\_lr\_warmup1\_00E-01} & 80.7 \\
        & \multirow{2}{*}{General} & \texttt{mlfoundations-dev/mistral\_7b\_0-3\_oh-} \newline \texttt{dcft-v3.1-gpt-4o-mini} & 69.6 \\
        \bottomrule
    \end{tabular}
    }
    \label{tab:mistral_gems}
\end{table*}

\clearpage

\begin{table*}[t!]
    \caption{\textbf{\textit{Qwen-7B Hidden Gems.}}}
    \vspace{0.2cm}
    \centering

    {
    \small
    \setlength{\tabcolsep}{8pt}
    \renewcommand{\arraystretch}{1.05}
    
    \begin{tabular}{ccccc}
        \textbf{Tree} & \textbf{Task} & \textbf{Best Found Performer} & \textbf{Accuracy} \\
        \toprule
        \multirow{3}{*}{\textbf{Qwen-7B}} & Coding & \texttt{mlx-community/Josiefied-Qwen2.5-Coder-7B-Instruct-abliterated-v1} & 82.1 \\
        & Math & \texttt{zwhe99/DeepMath-Zero-Math-7B} & 92.1 \\
        & General & \texttt{langfeng01/GiGPO-Qwen2.5-7B-Instruct-ALFWorld} & 79.1 \\
        \bottomrule
    \end{tabular}
    }
    \label{tab:qwen7b_gems}
\end{table*}

\end{document}